\documentclass[11pt,twoside]{article}

\usepackage{epsfig,amsfonts,color}
\usepackage{amsmath}
\usepackage{amssymb, palatino, geometry,url}
\usepackage[noresetcount,lined,boxed]{algorithm2e} % ... for algorithms

\usepackage[colorlinks=true,linkcolor=blue,citecolor=blue,urlcolor=blue]{hyperref}
\usepackage{amsmath}
\allowdisplaybreaks

\usepackage{fancyhdr}
\newcommand{\be}{\begin{equation}}
\newcommand{\ee}{\end{equation}}
\newcommand{\bea}{\begin{eqnarray}}
\newcommand{\eea}{\end{eqnarray}}

\newcommand{\bvec}{\left(\begin{array}{c}}
\newcommand{\evec}{\end{array}\right)}
\newcommand{\bsub}{\begin{subequations}}
\newcommand{\esub}{\end{subequations}}

\usepackage{lineno}
\usepackage{pdflscape}
\usepackage{subfigure}
\usepackage{algorithm2e}

\begin{document}

\title{New Paradigms for Exploiting Parallel Experiments in Bayesian Optimization}

\author{Leonardo D. Gonz\'alez and Victor M. Zavala\thanks{Corresponding Author: victor.zavala@wisc.edu.}\\
 \\
  {\small Department of Chemical and Biological Engineering}\\
 {\small \;University of Wisconsin-Madison, 1415 Engineering Dr, Madison, WI 53706, USA}}
 \date{}
\maketitle

\begin{abstract}
Bayesian optimization (BO) is one of the most effective methods for closed-loop experimental design and black-box optimization. However, a key limitation of BO is that it is an inherently sequential algorithm (one experiment is proposed per round) and thus cannot directly exploit high-throughput (parallel) experiments. Diverse modifications to the BO framework have been proposed in the literature to enable exploitation of parallel experiments but such approaches are limited in the degree of parallelization that they can achieve and can lead to redundant experiments (thus wasting resources and potentially compromising performance). In this work, we present new parallel BO paradigms that exploit the structure of the system to partition the design space. Specifically, we propose an approach that partitions the design space by following the level sets of the performance function and an approach that exploits partially separable structures of the performance function found. We conduct extensive numerical experiments using a reactor case study to benchmark the effectiveness of these approaches against a variety of state-of-the-art parallel algorithms reported in the literature. Our computational results show that our approaches significantly reduce the required search time and increase the probability of finding a global (rather than local) solution. 
\end{abstract}


%%%%%%%%%%%%%%%%%%%%%%%%%%%%%%%%%%%%%%%%%%

\section{Introduction}

The use of high-throughput experimental (HTE) platforms is accelerating scientific discovery in diverse fields such as catalysis \cite{nguyen2020htecatalysis}, pharmaceuticals \cite{mennen2019htepharma}, synthetic biology \cite{smanski2016htebio}, and chemical engineering \cite{selekman2017htecheme}. Such platforms permit large numbers of experiments to be executed in parallel, sometimes automatically; this enables the exploration of wider design spaces, reduces time to discovery, and can potentially decrease the use of resources. However, due to the large number of design variables involved, determining optimal conditions manually is often infeasible. As a result, HTE platforms rely on the use of design of experiments (DoE) algorithms, which aim to systematically explore the design space. 
\\

Screening is a simple DoE approach in which experiments are performed at points on a discretized grid of the design space \cite{shevlin2017htechem}; this approach is intuitive but does not scale well with the number of design variables and can ultimately lead to significant waste of resources (conduct experiments that do not provide significant information). The central aim of advanced DoE approaches is to maximize the value provided by each experiment and ultimately reduce the number of experiments and resources used (e.g., experiment time).  The value of an experiment is usually measured either by information content (e.g., reduces model uncertainty) or if it results in a desirable outcome (e.g., improves an economic objective) \cite{box2005doe}. A widely used DoE approach that aims to tackle this problem is response surface methodology or RSM \cite{box1951rsm}; this approach is generally sample-efficient (requires few experiments) but uses second-degree polynomial surrogate models that can fail to accurately capture system trends. In addition, parameters used in the RSM surrogate model are subject to uncertainty and this uncertainty is not resolved via further experiments \cite{jones2001rsm} (i.e., RSM is an open-loop DoE technique). 
\\

Another powerful approach to DoE that aims to maximize value of experiments is Bayesian experimental design \cite{chaloner1995bayesian}. Recently, the machine learning (ML) community has been using variants of this paradigm to conduct closed-loop experimental design \cite{ferguson2022doe}. One of the most effective variations of this paradigm is the Bayesian optimization (BO) algorithm \cite{biswas2021mobo};  BO has been shown to be sample-efficient and scalable (requires minimal experiments and can explore large design spaces) \cite{snoek2015scalable}. BO is widely used in applications such as experimental design, hyper-parameter tuning, and reinforcement learning. Of particular interest is the flexibility of the BO paradigm as it is capable of accommodating both continuous and discrete (e.g., categorical) design variables as well as constraints  (which help encode domain knowledge and restrict the design space) \cite{brochu2010tutorial}. Additionally, BO uses probabilistic surrogate models (e.g. Gaussian process models) which greatly facilitates the quantification of uncertainty and information in different regions of the design space \cite{garnett2021bayesian}; this feature is particularly useful in guiding experiments where information gain can be as important as performance. BO can also be tuned to emphasize exploration (by sampling regions with high uncertainty) over exploitation (by sampling from regions with high economic performance) \cite{mockus2012bayesian}; this trade-off is achieved by tuning the so-called acquisition function (AF), which is a composite function that captures uncertainty and performance. 
\\

A fundamental caveat of BO is that it is inherently a sequential algorithm (samples a single point in the design space at each iteration), limiting its ability to exploit HTE platforms. Modifications to the BO algorithm have been proposed in the literature to overcome these limitations \cite{mack2007surrogate, desautels2014bucb, marmin2015qei}. Relevant variants include Hyperspace partitioning \cite{young2018hyperspace}, \textcolor{black}{batch Bayesian optimization} \cite{wilson2017qBO}, NxMCMC \cite{snoek2012nmc}, and AF optimization over a set of exploratory designs \cite{hutter2012expw}. These parallel BO approaches have been shown to perform better than sequential BO in terms of search time \cite{young2020hyper}; however, these approaches are limited in the degree of parallelization that can be achieved and can lead to redundant experiments, thus wasting resources and potentially getting trapped in local solutions. 
\\

\begin{figure}[!htp]
	\begin{center}
		\includegraphics[width=5in]{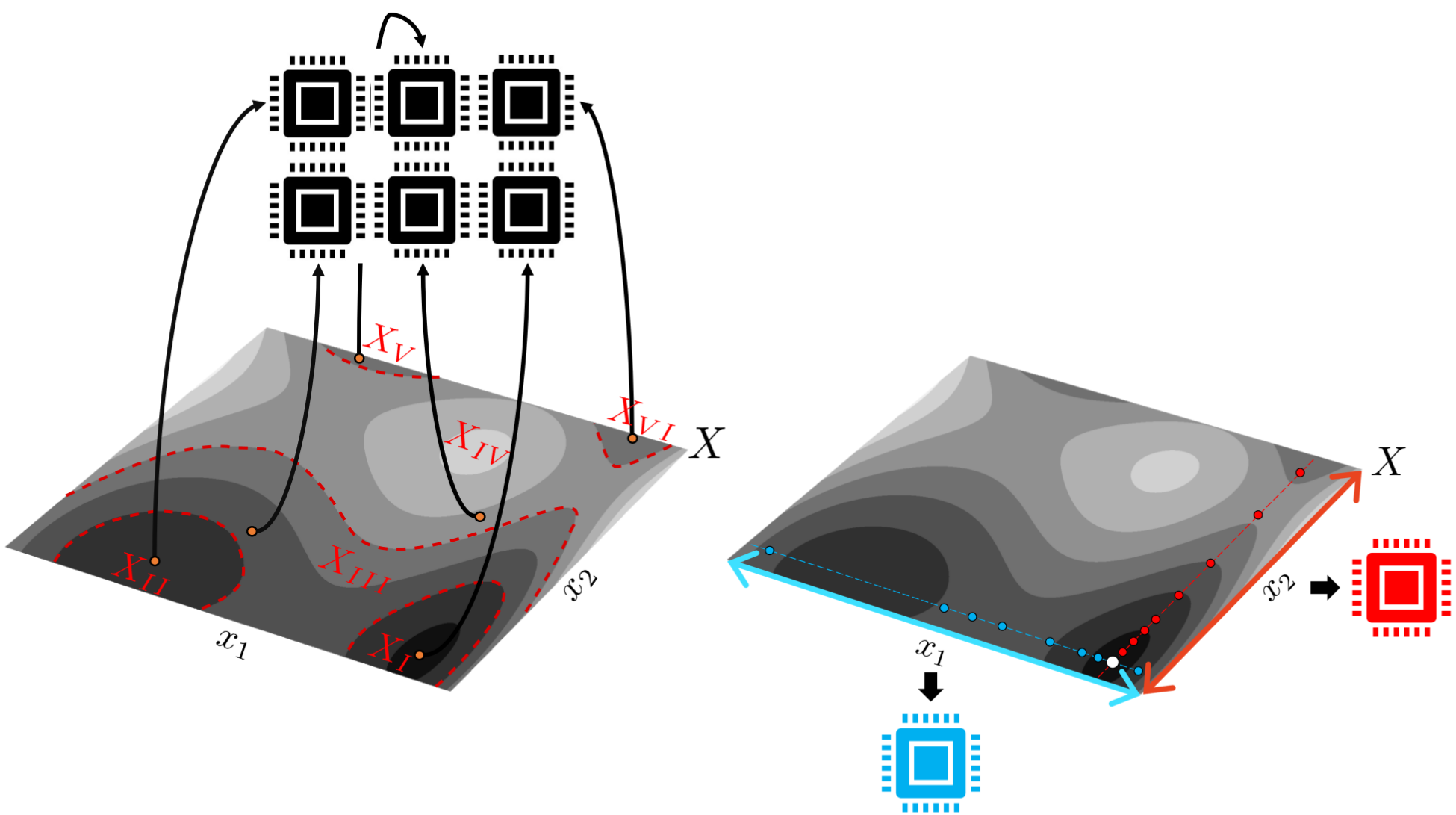}
		\caption{Schematic of proposed BO parallelization paradigms using level set partitioning (left) and variable partitioning (right).}
		\label{fig:Part_Algos}
	\end{center}
\end{figure}

In this work, we propose a set of new parallel BO paradigms that exploit the structure of the system in order to guide partitioning of the design space (see Figure \ref{fig:Part_Algos}).  Our first approach, which we call {\em level-set partitioning}, decomposes the design space by following the level sets of the performance function. Because the performance function cannot be evaluated (it is unknown), a key feature of this approach is that it leverages a reference function (which can be a low-fidelity model or a physics model) to approximate the level sets and guide the partitioning. Our second approach, called {\em variable partitioning}, partitions the design space by exploiting partially separable structures that typically result when a system is composed of multiple subsystems (e.g., a chemical process is composed of multiple units). We benchmark the performance of our approaches over sequential BO and state-of-the-art parallel BO variants from the literature using a reactor system. Our results show that the proposed approaches can achieve significant reductions in search time; in addition, we observe improvements in performance values found and in search robustness (sensitivity to initial guess). 

%%%%%%%%%%%%%%%%%%%%%%%%%%%%%%%%%%%%%%%%%%

\section{Sequential Bayesian Optimization}

\subsection{Standard BO \textcolor{black}{(S-BO)}}

The aim of closed-loop DoE is to identify experimental inputs that achieve optimal performance; we cast this as the optimization problem:
\begin{subequations}\label{DoE_goal}
    \begin{gather}
        \min_{x}~~f(x)\\
         \textrm{s.t.}~~x\in{X}
    \end{gather}
\end{subequations}
where $f: X\to \mathbb{R}$ is a scalar {\em performance function}, $x\in X\subseteq \mathbb{R}^d$ is a given experiment (trial) point; and $X$ is the experimental design space (of dimension $d$). Generally, an explicit relationship between the design variables $x$ and performance function $f$ is not known a priori and motivates the need to evaluate the performance at a selected set of trial points to build a surrogate model of the performance function that can be used to optimize the system. 
\\

The open-loop DoE approach most commonly used in HTE platforms is screening; this is a grid-search method that discretizes the design space $X$ into a set of experiments $x_k\in X,\; k\in \mathcal{K}:=\{1,...,K\}$ (we denote this set compactly as $x_{\mathcal{K}}$); the performance of the system is then evaluated (potentially in parallel) at these points to obtain $f_{\mathcal{K}}$ and the experiments that achieve the best performance are selected. This screening approach provides good exploratory capabilities, but it is not scalable in the sense that the number of trial experiments needed to cover the design space grows exponentially with the number of design variables $d$ and with the width of the space $X$.  Moreover, this approach cannot be guaranteed to find an optimal solution. 
\\

To derive our closed-loop DoE approach based on S-BO, we assume that we have a set of experimental data $\mathcal{D}^\ell=\{x_\mathcal{K}^\ell, f_\mathcal{K}^\ell\}$ at the initial iteration $\ell=1$. We use this data to develop a probabilistic surrogate model $\hat{f}^\ell:X\to \mathbb{R}$ (a Gaussian process - GP) that approximates the performance function $f$ over the design space $X$. \textcolor{black}{The GP generates this approximation by first constructing a prior over $f(x_{1:n})$ of the form $f(x_{1:n})\sim\mathcal{N}(\textbf{m}(x), \textbf{K}(x, x^{\prime}))$. Here $\textbf{m}(x)\in\mathbb{R}^d$ is the prior mean function and is commonly set equal to $\textbf{0}$. The prior covariance matrix, $\textbf{K}(x, x^{\prime})\in\mathbb{R}^{d\times d}$, is constructed using a kernel function, $h(x, x^{\prime})$, such that $\textbf{K}_{i,j}=h(x_i, x_j)$. There exists a large selection of kernel functions (e.g, rotational quadratic, squared exponential, M\'atern), and determining which to use is largely dependent on identifying a choice for $h(x,x^{\prime})$ that will fit the generated data well. The GP then uses these elements to construct a predictive posterior distribution at a new experimental point $x$ that generates an estimate $\hat{f}^{\ell}(x)$ of the value of the performance function at $x$. This estimate can be shown to be Gaussian random variable $\hat{f}^\ell(x)\sim \mathcal{N}(\mu_f^\ell(x),\sigma_f^\ell(x))$ with mean and variance:} 
\begin{subequations}\label{GP_moments}
    \begin{align}
        \mu_f^\ell(x) &:= \textbf{K}(x, x_\mathcal{K}^\ell)\textbf{K}(x_\mathcal{K}^\ell, x_\mathcal{K}^\ell)^{-1}f_\mathcal{K}^\ell,\; x\in X\\
        \sigma_f^\ell(x) &:= \textbf{K}(x, x)-\textbf{K}(x, x_\mathcal{K}^\ell)^T\textbf{K}(x_\mathcal{K}^\ell, x_\mathcal{K}^\ell)^{-1}\textbf{K}(x_\mathcal{K}^\ell, x), \; x\in X.
    \end{align}
\end{subequations}
The mean and variance of the prediction capture the expected performance and its uncertainty/information; these measures are used to construct an acquisition function (AF),  which is used to guide the selection of the next experiment. Here, we use an AF of the form:  
\begin{equation}\label{LCB}
    AF_f^\ell(x; \kappa ) = \mu_f^\ell(x)-\kappa\cdot \sigma_f^\ell(x), \; x\in X
\end{equation}
where $\kappa \in \mathbb{R}_+$ is a hyperparameter that prioritizes exploration (information) over exploitation (performance). The next experiment, $x^{\ell+1}\in X$, is obtained by solving the AF optimization problem:
\begin{subequations}\label{LCB_min}
    \begin{gather}
       x^{\ell+1}\leftarrow \mathop{\textrm{argmin}}_{x}~~AF_f^\ell(x; \kappa)\\
        \textrm{s.t.}~~x\in{X}
    \end{gather}
\end{subequations}
Once the experiment $x^{\ell+1}$ has been computed, the system performance is evaluated at this point (by running the experiment) to obtain the new data point $\{ x^{\ell+1}, f(x^{\ell+1})\}$; this point is added to the data collection $\mathcal{D}^{\ell+1}\leftarrow \mathcal{D}^\ell \cup  \{ x^{\ell+1}, f(x^{\ell+1})\}$. The GP model is re-trained using this new data collection to obtain $\hat{f}^{\ell+1}(x)\sim \mathcal{N}(\mu_f^{\ell+1}(x),\sigma_f^{\ell+1}(x))$ and the acquisition function is updated to $AF_f^{\ell+1}$ and minimized to obtain a new design $x^{\ell+2}$. This process is repeated over multiple cycles/iterations $\ell=1,2,...,L$ until a satisfactory performance is obtained (or no additional improvement in performance is observed). We highlight that the standard BO algorithm (which we refer to as S-BO) does not have a natural stopping criterion and can get stuck in a local solution (as opposed to finding a global solution). \textcolor{black}{The pseudocode for S-BO can be found in Algorithm \ref{alg:BO}.}

\begin{figure}[hbt!]
	\begin{center}
		\includegraphics[width=4.5in]{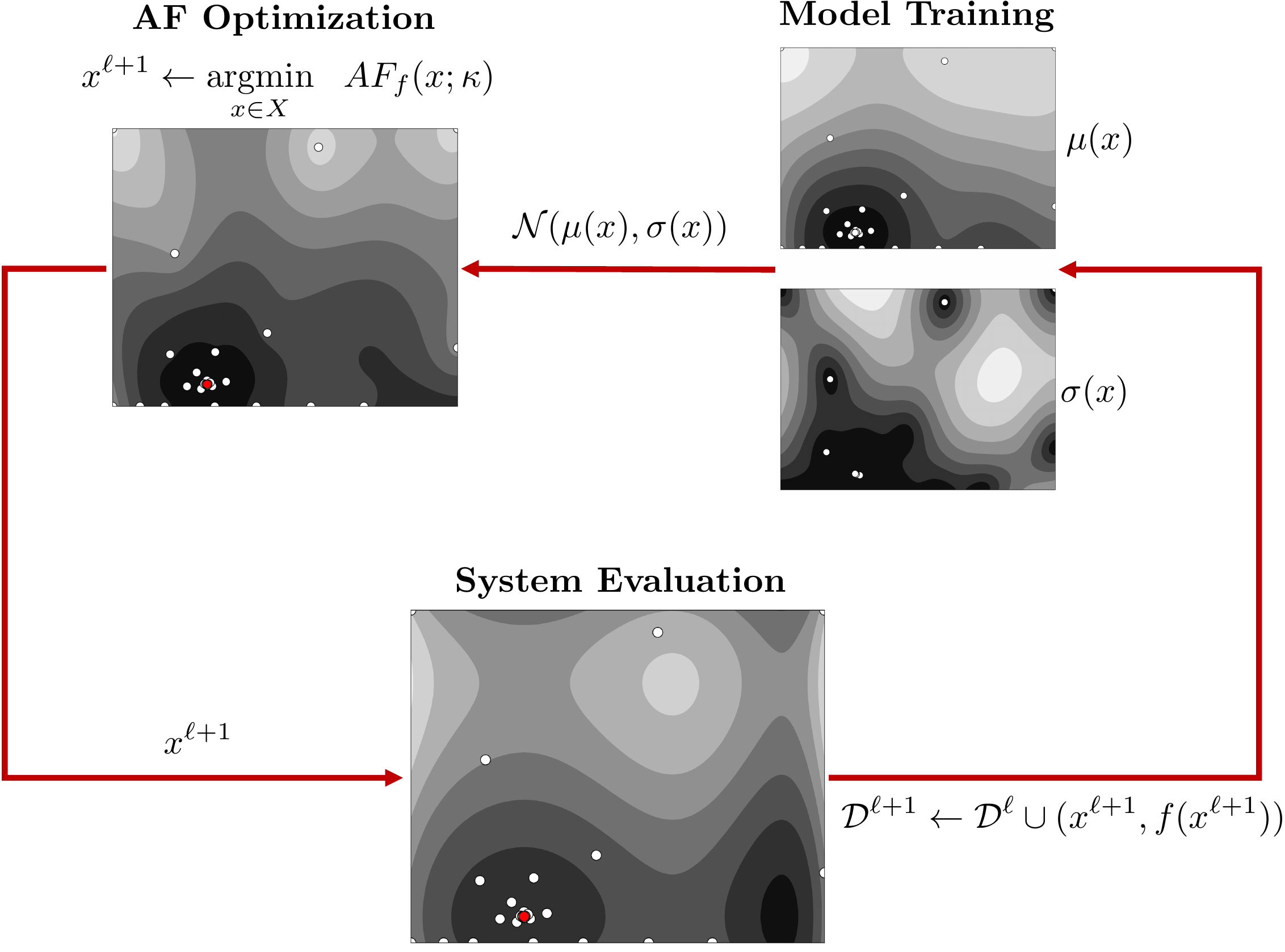}
		\caption{Workflow of a standard Bayesian optimization (S-BO) framework. A new experiment $x^{\ell+1}$ is obtained from the optimization of the acquisition function. The system performance is evaluated at the experiment and the collected data is added to the database. The database is used to re-train the probabilistic GP model (which captures performance and uncertainty). The GP model is used to construct the acquisition function, which balances performance and uncertainty.}
		\label{fig:BO_Algo}
	\end{center}
\end{figure}

\RestyleAlgo{ruled}
\begin{algorithm}[hbt!]
\caption{Standard Bayesian Optimization \textcolor{black}{(S-BO)}}\label{alg:BO}
Given $\kappa$, $L$, and $\mathcal{D}^{\ell}$\;
Train GP $\hat{f}^\ell$ using initial dataset $\mathcal{D}^{\ell}$ and obtain $AF_f^\ell$\;
\For{$\ell=1, 2,..., L$}{
Compute experiment $x^{\ell+1} \gets \mathop{\textrm{argmin}}_{x}{AF^{\ell}_f(x;\kappa)}$ s.t. $x\in X$\;
Evaluate performance at $x^{\ell+1}$ to obtain $f^{\ell+1}$\;
Update dataset $\mathcal{D}^{\ell+1}\gets \mathcal{D}^{\ell}\cup  \left\{x^{\ell+1}, f^{\ell+1}\right\}$\;
Train GP using $\mathcal{D}^{\ell+1}$ to obtain $\hat{f}^{\ell+1}$ and $AF_f^{\ell+1}$\;
}
\end{algorithm}

\subsection{Reference-Based BO (Ref-BO)}

\textcolor{black}{Recent work has shown that allowing the BO algorithm to exploit available preexisting information can improve its performance. One approach found in \cite{paulson2021cobalt}, for example, exploits the fact that the performance function is usually a known composite function that can be expressed as $f(y(x))$ where a set of intermediate variables, $y(x)$, are the unknowns. As a result, the performance function can be optimized using derivative-based methods, allowing one to set targets for the various $y(x)$ which are modeled by the BO algorithm to determine the appropriate input $x$ values. Another approach uses a low fidelity estimate of $f(x)$ to identify promising regions in $X$ and then increases the fidelity of the estimate while sampling from this reduced space; this process is then repeated iteratively. This method, known as multi-fidelity BO (MFBO), gradually zeros in on an optimal region, reducing the number of experiments that have to be performed with the real system \cite{farshud2021mulfi, kandasamy2017mulfi, wu2020mulfi}. The approach we have opted to use, which we refer to as reference-based BO or Ref-BO, is based on the framework presented in \cite{lu2021hvac}, and is similar to MFBO. Specifically, Ref-BO initializes the BO algorithm with an approximation of $f(x)$ in order to highlight promising regions in the design space where the solution might be located; BO can then focus sampling in such regions from the start and avoid unnecessary experiments. We refer to this initial approximation as the reference model, and it can be obtained through various means (e.g.,  physics models, empirical correlations, or low-fidelity simulators). Unlike MFBO, Ref-BO always samples from the real system and does not use the low-fidelity model to restrict sampling to any one region of the design space; additionally, the reference model is not modified after it has been loaded into the algorithm. As a result, the Ref-BO algorithm is simple to implement. Once the reference model has been obtained, only a few minor modifications to S-BO are required without the need to make any additional assumptions or set additional hyperparameters beyond those associated with S-BO.} 
\\

To incorporate a reference model in the BO search, we reformulate the optimization problem in \eqref{DoE_goal} as:
\begin{subequations}
	\begin{gather}
		\min_{x}\ \ g(x)+\varepsilon(x)\\
		\textrm{s.t.}\ \ x\in X
	\end{gather} \label{eq: BO_ReferenceModel}
\end{subequations}
where $g:X\rightarrow \mathbb{R} $ is the reference model and $\varepsilon:X\rightarrow \mathbb{R}$ is the residual or error model satisfying $\varepsilon(x)=f(x)-g(x),\; x\in X$. The reference model $g$ is assumed to be a deterministic function that captures features of the performance function $f$. We also assume that, during the BO search, the reference model is fixed and unaffected by new data collected. 

\begin{figure}[!htp]
	\begin{center}
		\includegraphics[width=5in]{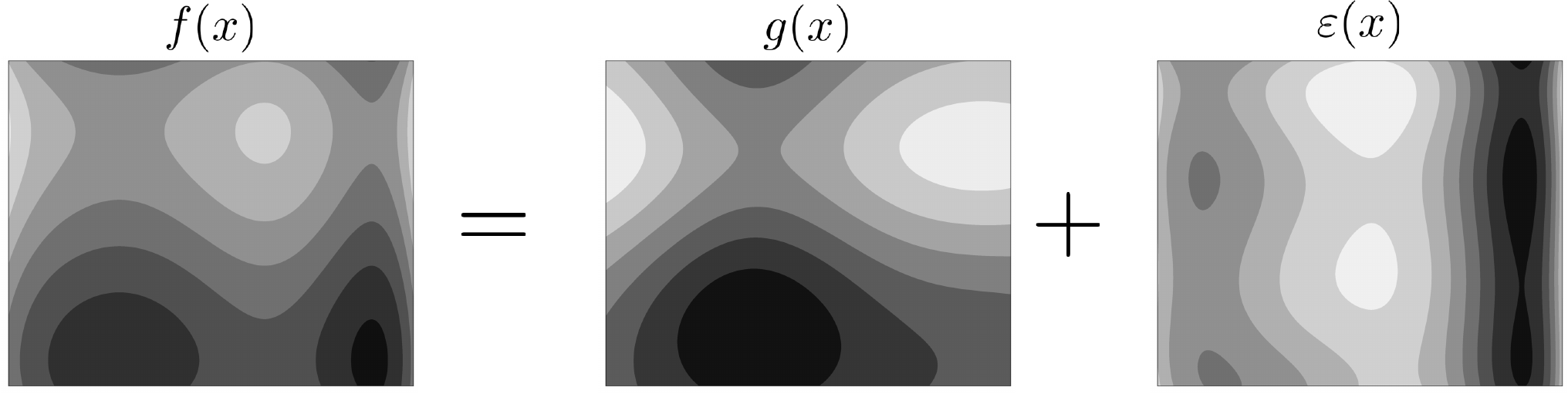}
		\caption{Reference-based BO decomposes the performance function $f$ into a reference model $g$ and a residual model $\varepsilon$. The coarse features of $f$ are often captured by $g$ and thus the residual $\varepsilon$ is typically easier to learn using a GP model.}
		\label{fig:REF_BO}
	\end{center}
\end{figure}

The form of the residual model $\varepsilon$ is not known (because $f$ is not known) and thus a surrogate needs to be built from experimental data. Given a set of data $\mathcal{D}_\varepsilon^\ell :=\{x_\mathcal{K}^\ell,\varepsilon_\mathcal{K}^\ell\}$, with $\varepsilon_\mathcal{K}^\ell:=\{f(x_k^\ell)-g(x_k^\ell)\}_{k\in \mathcal{K}}$, we construct a GP model for the residual; the prediction of the residual at a new point $x$ is the Gaussian $\hat{\varepsilon}^\ell(x)\sim \mathcal{N}(\mu_\varepsilon^\ell(x),\sigma_\varepsilon^\ell(x))$ with: 
\begin{subequations}\label{GP_moments}
    \begin{align}
        \mu_\varepsilon^\ell(x) &:= \textbf{K}(x, x_\mathcal{K}^\ell)\textbf{K}(x_\mathcal{K}^\ell, x_\mathcal{K}^\ell)^{-1}\varepsilon_\mathcal{K}^\ell,\; x\in X\\
        \sigma_\varepsilon^\ell(x) &:= \textbf{K}(x, x)-\textbf{K}(x, x_\mathcal{K}^\ell)^T\textbf{K}(x_\mathcal{K}^\ell, x_\mathcal{K}^\ell)^{-1}\textbf{K}(x_\mathcal{K}^\ell, x),\; x\in X.
    \end{align}
\end{subequations}
We build the surrogate $\hat{f}=g+\hat{\varepsilon}$ of the performance function based on the surrogate of the residual; because $g$ is assumed to be deterministic, we have that:  
\begin{align}
		\hat{f}^\ell(x)\sim\mathcal{N}(g(x)+\mu_\varepsilon^\ell(x),\sigma_\varepsilon^\ell(x)^2)\label{eq: BO_Reference_GP}
\end{align}
This indicates that we need to modify the AF used in S-BO as:
\begin{equation}
	AF_\varepsilon^\ell(x;\kappa)=(g(x)+\mu_\varepsilon^\ell(x))-\kappa \cdot \sigma_\varepsilon^\ell(x),\; x\in X \label{ModifiedAF}
\end{equation}
This AF is minimized to obtain the next experiment and residual evaluation $\{x^{\ell+1},\varepsilon(x^{\ell+1})\}$, with $\varepsilon(x^{\ell+1})=g(x^{\ell+1})-f(x^{\ell+1})$, and we use this data point to update the data set $\mathcal{D}_\varepsilon^{\ell+1}$. We can thus see that BO with a reference model is analogous to standard BO. \textcolor{black}{The general Ref-BO framework is presented in Algorithm \ref{alg:Ref-BO}.}

\RestyleAlgo{ruled}
\begin{algorithm}[hbt!]
\caption{Reference-Based Bayesian Optimization (Ref-BO)}\label{alg:Ref-BO}
Given reference model $g$, $\kappa$, $L$, and $\mathcal{D}_\varepsilon^{\ell}$\; 
Train GP $\hat{\varepsilon}^\ell$ using $\mathcal{D}_\varepsilon^{\ell}$ and obtain $AF_\varepsilon^\ell$\;
\For{$\ell=1, 2,..., L$}{
Compute experiment $x^{\ell+1} \gets \mathop{\textrm{argmin}}_{x}{AF^{\ell}_\varepsilon(x;\kappa)}$ s.t. $x\in X$\;
Evaluate performance at $x^{\ell+1}$ to obtain $f^{\ell+1}$ and residual $\varepsilon^{\ell+1}$\;
Update dataset $\mathcal{D}_\varepsilon^{\ell+1}\gets \mathcal{D}_\varepsilon^{\ell}\cup  \left\{x^{\ell+1}, \varepsilon^{\ell+1}\right\}$\;
Train GP using $\mathcal{D}_\varepsilon^{\ell+1}$ to obtain $\hat{\varepsilon}^{\ell+1}$\;
}
\end{algorithm}

\section{Parallel Bayesian Optimization}

The S-BO and Ref-BO frameworks discussed are sample efficient (typically require few experiments to identify optimal performance) but they are inherently sequential. Several approaches for proposing multiple experiments per cycle have been developed, each with varying degrees of complexity and sample efficiency. These parallel BO variants can grouped into four main parallelization paradigms: AF optimization over a set of hyperparameters, design space partitioning, fantasy sampling, \textcolor{black}{and AF optimization over a batch of points.} The most used approach is the NxMCMC method, which falls under the fantasy sampling paradigm, and is used in popular BO packages such as Spearmint \cite{spearmint2012}. We now proceed to discuss the specifics of different existing algorithms that are based on these parallelization paradigms. 
\\

\subsection{Hyperparameter Sampling Algorithm (HP-BO)}

The hyperparameter sampling algorithm (which we refer to as HP-BO) identifies a new {\em batch} of experiments $x_\mathcal{K}^{\ell+1}$ by optimizing the acquisition function $AF^\ell(x;\kappa_k)$ using $K$ different values of the exploratory hyperparameter $\kappa_k,\,k\in \mathcal{K}$.  In other words, we obtain the new experiments by solving:
\begin{subequations}\label{LCB_sample_hyper}
    \begin{gather}
        x_{k}^{\ell+1}\leftarrow \mathop{\textrm{argmin}}_{x}~~{AF}^\ell_f(x; \kappa_k)\\
        \textrm{s.t.}~~x\in{X}
    \end{gather}
\end{subequations}
for $k\in \mathcal{K}$. The hyperparameter values $\kappa_k,\ k\in \mathcal{K}$, can be selected manually or sampled from a distribution. In the approach that we consider here, we generate the values by sampling from an exponential distribution $\kappa\sim \mathcal{E}(\lambda)$ with rate parameter $\lambda=1$ as shown in \cite{hutter2012expw}.  Once the batch of experiments has been determined, we can evaluate their performance (in parallel) to obtain $f_k^{\ell+1}=f(x_k^{\ell+1}),\, k\in \mathcal{K}$ and update the data $\mathcal{D}^{\ell+1}\gets \mathcal{D}^{\ell}\cup  \left\{x_\mathcal{K}^{\ell+1}, f_\mathcal{K}^{\ell+1}\right\}$. The updated data is then used to train a new GP $\hat{f}^{\ell+1}$, which is used to form a new acquisition  $AF^{\ell+1}_f$, and to compute a next batch of experiments $x_{\mathcal{K}}^{\ell+2}$. The process is repeated over multiple cycles. \textcolor{black}{The pseudocode for implementing HP-BO is shown in Algorithm \ref{alg:HP-BO}.}

\begin{figure}[!htp]
	\begin{center}
		\includegraphics[width=5.0in]{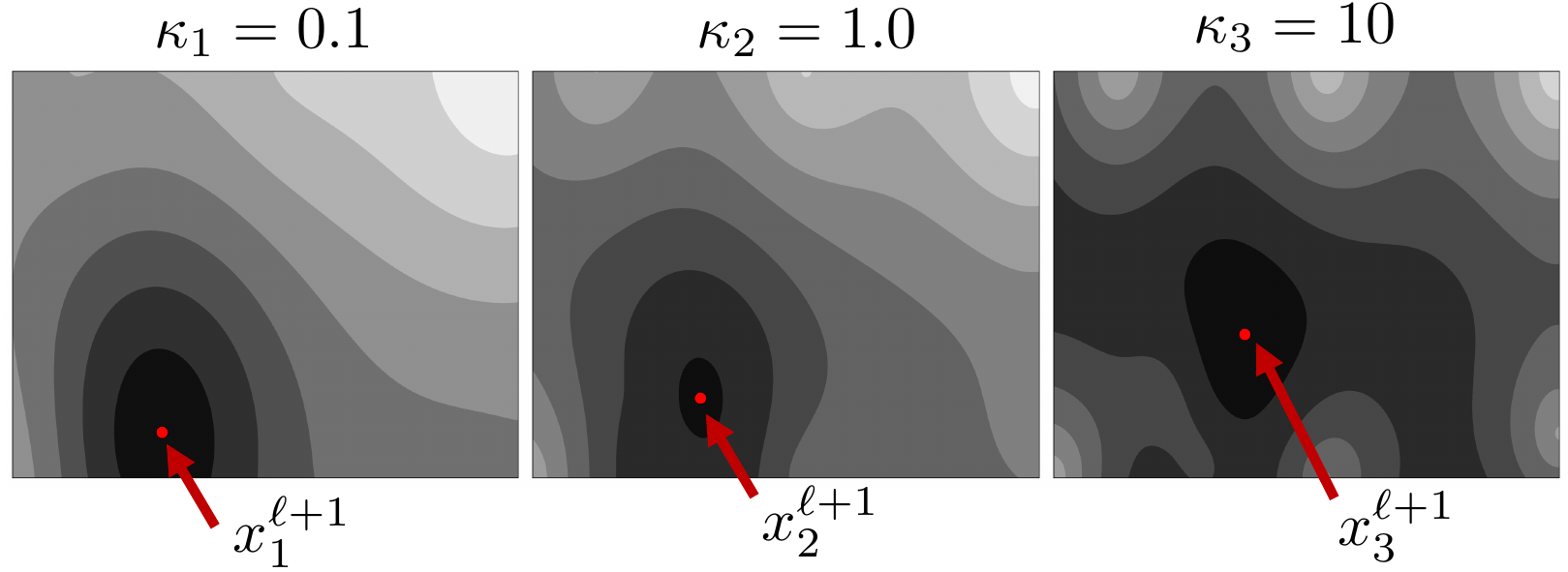}
		\caption{HP-BO optimizes the AF for a set of hyperparameters $\kappa_k,\ k\in \mathcal{K}$ to obtain experiments $x_k,\ k\in \mathcal{K}$ that can be evaluated in parallel. Here, we show an example with $K=3$.}
		\label{fig:HYPP_BO}
	\end{center}
\end{figure}

The main advantages of HP-BO are that it is easy to implement, that it is \textcolor{black}{highly} parallelizable, and that it allows for the selection of experiments under various exploration and exploitation settings (eliminating the need for tuning $\kappa$). The effect of the hyperparameter $\kappa$ on the AF is highlighted in Figure \ref{fig:HYPP_BO}. However, in this approach it is not possible to prevent the proposal of redundant experiments and, as the algorithm converges, the suggested experiments can begin to cluster in a region of low uncertainty (this can cause the algorithm to get trapped at local solutions). The HP-BO algorithm can be easily be extended to incorporate a reference function $g$; in this case, the GP learns the residual instead of the performance function. 

\begin{algorithm}[!htp]
\caption{Hyperparameter Sampling BO (HP-BO)}\label{alg:HP-BO}
Given $\kappa_k,\,k\in \mathcal{K}$, $L$, and $\mathcal{D}^{\ell}$\;
Train GP $\hat{f}^\ell$ using $\mathcal{D}^{\ell}$ and obtain $AF_f^\ell(x; \kappa_k),\,k\in \mathcal{K}$\;
\For{$\ell=1, 2,..., L$}{
\For{$k\in \mathcal{K}$}{
Compute experiment $x_k^{\ell+1}\gets\mathop{\textrm{argmin}}_{x}{AF^{\ell}_f(x;\kappa_k)}$ s.t. $x\in X$\;
Evaluate performance at $x_{k}^{\ell+1}$ to obtain $f_k^{\ell+1}$\;
}
Update data $\mathcal{D}^{\ell+1}\gets \mathcal{D}^{\ell}\cup  \left\{x_\mathcal{K}^{\ell+1}, f_\mathcal{K}^{\ell+1}\right\}$\;
Train GP using $\mathcal{D}^{\ell+1}$ to obtain $\hat{f}^{\ell+1}$\;}
\end{algorithm}

\subsection{HyperSpace Partitioning Algorithm (HS-BO)}

\begin{figure}[!htp]
	\begin{center}
		\includegraphics[height=1.5in]{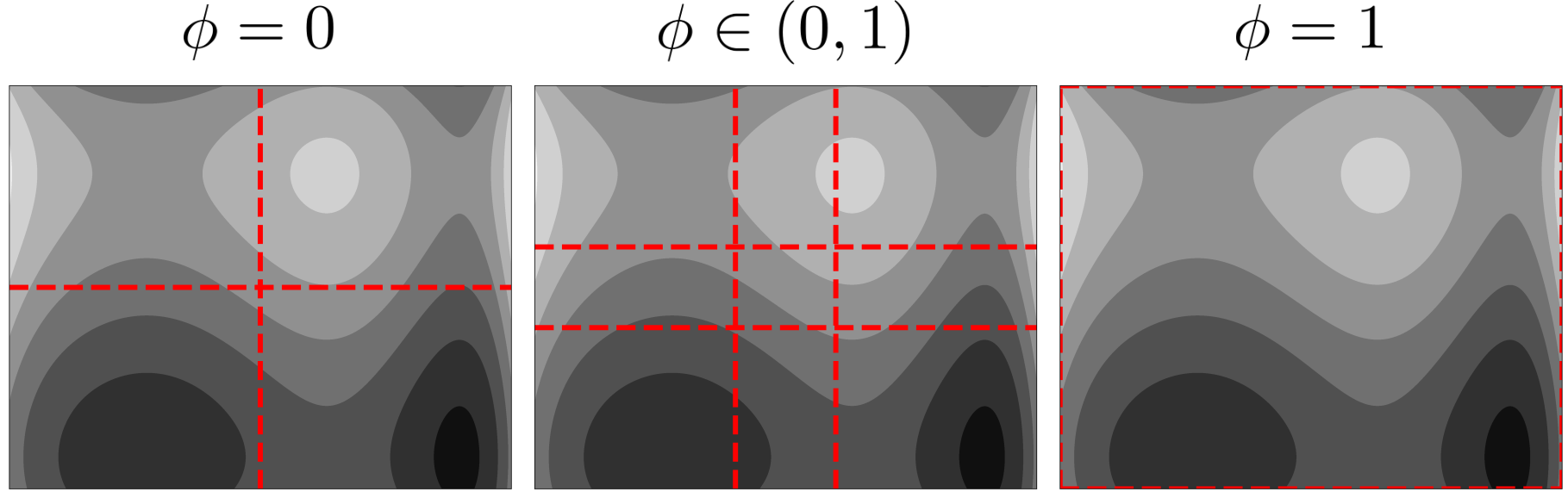}
		\caption{HS-BO partitions the domain $X$ into $K=2^d$ subdomains and runs a separate instance of S-BO within each partition. A hyperparameter $\phi$ is introduced to define the degree of overlap in the partitions (the overlapping region aims to share information across subdomains). When $\phi=0$ there is no overlap between the partitions and when $\phi=1$ we have that all partitions are the entire domain $X$.}
		\label{fig:HYPS_BO}
	\end{center}
\end{figure}

The HyperSpace partitioning algorithm (which we refer to as HS-BO) was presented in \cite{young2018hyperspace}; this parallelizes BO by partitioning the design space $X$ into $K$ equally-sized blocks $X_k\subseteq X,\,k\in \mathcal{K}$. Importantly, this approach does not use a surrogate GP model over the entire design space; instead, a separate GP model is constructed at each partition $X_k$ and is updated using only information collected within this partition. Specifically, each partition $k\in \mathcal{K}$ builds a GP $\hat{f}^\ell_k\sim \mathcal{N}(\mu_{f,k}^\ell(x),\sigma_{f,k}^\ell(x))$ that is used to construct an acquisition function $AF_{f,k}^\ell(x; \kappa)$. With this, we can obtain a set of new experiments by solving the following subproblems:
\begin{subequations}\label{LCB_sample_hyper}
    \begin{gather}
        x_{k}^{\ell+1}\leftarrow \mathop{\textrm{argmin}}_{x}~~{AF}^\ell_{f,k}(x; \kappa)\\
        \textrm{s.t.}~~x\in{X}_k
    \end{gather}
\end{subequations}
for $k\in \mathcal{K}$.  The domain partitions can also be constructed to have a certain degree of overlap; specifically, an overlap hyperparameter $\phi\in[0, 1]$ is introduced to allow the partitions to share a fraction of the design space. A value of $\phi=0$ indicates that the partitions $X_k$ are completely separate, while a value of $\phi=1$ indicates that $X_k=X$ for $k\in\mathcal{K}$ (the partitions are copies of the full design space); this is shown in Figure \ref{fig:HYPS_BO}. The overlap hyperparameter provides a communication window, allowing the GP model of a given partition to observe system behavior beyond its prescribed partition (share information with other partitions).  This, however, introduces a fundamental trade-off; from a parallelization perspective it is desirable that $\phi$ is small, but from convergence perspective (e.g., reducing number of iterations) it might be desirable that $\phi$ is large.  A similar trade-off is observed as one decreases or increases the number of partitions $K$; as such, there is a complex interplay between the hyperparameters $K,\phi$, and these need to be tuned. In the implementation reported in  \cite{young2018hyperspace}, the number of partitions is set to $K=2^d$. Similar types of trade-offs have been observed in the context of overlapping decomposition approaches for optimization problems defined over graph domains  \cite{shin2020decentralized}. 

The HS-BO approach, \textcolor{black}{summarized in Algorithm \ref{alg:HS-BO}}, is easy to implement, is scalable to high-dimensional spaces, and enables the development of GP models for systems that may exhibit different behaviors at various regions of the design space (compared to using a single GP model that captures the entire design space). HS-BO also eliminates redundant sampling by forcing the algorithm to sample from distinct regions of the design space. This also results in a more thorough search, which improves the probability that the global solution will be located; domain partitioning is, in fact, a paradigm widely used in global optimization. A limitation of HS-BO is that the partitions are boxes of equal size (this can limit capturing complex shapes of the performance function); moreover, one needs to tune $K$ and $\phi$. In principle, it might be possible to extend this approach to account for automatic tuning and adaptive partitions, but this would require much more difficult implementations that carefully trade-off parallelization and convergence (this is left as a topic of future work).  The HS-BO approach can also be easily executed using a reference model by learning the residual instead of the performance function. 

\begin{algorithm}[!hbt]
\caption{HyperSpace Partitioning (HS-BO)}\label{alg:HS-BO}
Given $\kappa$, $K$, $L$, $\phi$, and $\mathcal{D}^{\ell}$\;
Partition $X$ into $X_k\subseteq X,\,k \in\mathcal{K}$ with overlap $\phi$\;
Split initial data into domains $\mathcal{D}^\ell_{k},\,k\in \mathcal{K}$\; 
\For{$k\in \mathcal{K}$}{
Train GP $\hat{f}_k$ in $X_k$ using $\mathcal{D}_k^{\ell}$\;}
\For{$\ell=1, 2,..., L$}{
\For{$k\in\mathcal{K}$}{
Compute experiment $x_k^{\ell+1}\gets\mathop{\textrm{argmin}}_{x}{AF^{\ell}_{f,k}(x;\kappa)}$ s.t. $x\in X_k$\;
Evaluate performance at $x_{k}^{\ell+1}$ to obtain $f_{k}^{\ell+1}$\;
Update data $\mathcal{D}_k^{\ell+1}\gets \mathcal{D}_k^{\ell}\cup  \left\{x_{k}^{\ell+1}, f_{k}^{\ell+1}\right\}$\;
Train GP using $\mathcal{D}_k^{\ell+1}$ to obtain $\hat{f}_k^{\ell+1}$\;}}
\end{algorithm}

\subsection{N$\times$MCMC Algorithm (MC-BO)}

The N$\times$MCMC (N times Markov Chain Monte Carlo) algorithm is a popular approach used for proposing multiple experiments \cite{shahriari2016BOreview}.  We refer to this approach simply as MC-BO. Assume that we currently have a set of experimental data $\mathcal{D}^\ell$; we use this to generate the GP $\hat{f}^\ell$, acquisition function $AF_{f}^{\ell}$, and to compute the next experiment $x_k^{\ell+1}$ for $k=1$. Our goal is now to obtain the remaining set of experiments $x_k^{\ell+1},\, k=2,...,K$ that we can use to evaluate performance. To do so, we consider a set of {\em fantasy} predictions obtained by generating $S$ samples from the GP $\hat{f}^{\ell}_s(x_k^{\ell+1}),\, s\in \mathcal{S}$.  The term {\em fantasy} alludes to the fact that the evaluation of performance is based on the GP (and not on the actual system). The fantasy data has the goal of creating an approximate AF; specifically, for each sample $s$, we generate a data point $\hat{\mathcal{D}}_{s}:=\{x^{\ell+1}_k, \hat{f}^{\ell}_s(x_k^{\ell+1})\}$ and this is appended to the existing dataset $\mathcal{D}^\ell\cup \hat{\mathcal{D}}_{s}$; this data is then used to obtain a GP $\hat{f}_s$ and associated acquisition function $\hat{AF}_{f,s}$. The AFs for all samples $s\in \mathcal{S}$ are collected and used to compute the mean AF: 
\begin{equation}\label{NxMC}
        \overline{AF}_f(x;\kappa):=\frac{1}{S}\sum_{s\in\mathcal{S}}\hat{AF}_{f,s}(x;\kappa),\; x\in X.
\end{equation}
The new experiment is then obtained by solving:
\begin{subequations}\label{LCB_sample}
    \begin{gather}
        x_{k+1}^{\ell+1}\leftarrow \mathop{\textrm{argmin}}_{x}~~\overline{AF}_f(x; \kappa)\\
        \textrm{s.t.}~~x\in{X}
    \end{gather}
\end{subequations}
To generate another experiment, we repeat the sampling process (using a new set of $S$ samples) and create a different mean acquisition function $\overline{AF}_f$, and we minimize this to obtain $x_{k+2}^{\ell+1}$. The sampling process is repeated until we have the full batch of new experiments $x_{\mathcal{K}}^{\ell+1}$. Once we have these, we evaluate the performance function at these points (in parallel) to obtain the dataset $\{x_{\mathcal{K}}^{\ell+1},f_{\mathcal{K}}^{\ell+1}\}$, which we append to the data collection  $\mathcal{D}^{\ell+1}\leftarrow \mathcal{D}^\ell\cup \{x_{\mathcal{K}}^{\ell+1},f_{\mathcal{K}}^{\ell+1}\}$. We use this data collection to re-train the GP of the performance function and repeat the process. \textcolor{black}{The framework for the MC-BO algorithm is presented in Algorithm \ref{alg:MC-BO}.}
\\

The MC-BO algorithm has proven to be an effective parallel extension of the BO algorithm; however, computing the mean AF requires significant computational time (as the GP model needs to be retrained continuously). This algorithm also has the tendency to propose experiments that are close in the design space, especially when it begins to converge. This does not necessarily pose an issue if the algorithm is converging to global solution; however, if the solution approached is local, this behavior can limit the ability of the algorithm to escape this region.  The MC-BO approach can be executed using a reference model by simply learning the residual instead of the performance function. 

\newpage 
\RestyleAlgo{ruled}
\begin{algorithm}[!ht]
\caption{N$\times$MCMC BO Algorithm (MC-BO)}\label{alg:MC-BO}
Given $\kappa$, $S$, $L$, and $\mathcal{D}^{\ell}$\;
Train GP $\hat{f}^\ell$ using initial dataset $\mathcal{D}^{\ell}$\;
\For{$\ell=1, 2,..., L$}{
Compute $x^{\ell+1}_k \gets \mathop{\textrm{argmin}}_{x}{AF^{\ell}_f(x;\kappa)}$ s.t. $x\in X$ for $k=1$\;
\For{$k=1,...,K-1$}{
\For{$s\in \mathcal{S}$}{
Generate fantasy data point $\hat{\mathcal{D}}_s=\left\{x_{k}^{\ell+1}, \hat{f}^{\ell}_s(x_k^{\ell+1})\right\}$\;
Use dataset $\mathcal{D}^{\ell}\cup\hat{\mathcal{D}}_s$ to train GP $\hat{f}_{s}$ and obtain $\hat{AF}_{f,s}(x;\kappa)$\;
}
Set $\overline{AF}_f(x; \kappa) \gets \frac{1}{S}\sum_{s\in \mathcal{S}}\hat{AF}_{f,s}(x;\kappa)$\;
Compute experiment $x_{k+1}^{\ell+1}\gets \mathop{\textrm{argmin}}_{x}{\overline{AF}_f(x;\kappa)}$ s.t. $x\in X$\;
}
\For{$k\in \mathcal{K}$}{
Evaluate performance at $x_{k}^{\ell+1}$ to obtain $f^{\ell+1}_{k}$\;
}
Update data $\mathcal{D}^{\ell+1}\gets \mathcal{D}^{\ell}\cup  \left\{x_\mathcal{K}^{\ell+1}, f_\mathcal{K}^{\ell+1}\right\}$\;
Train GP using $\mathcal{D}^{\ell+1}$ to obtain $\hat{f}^{\ell+1}$\;
}
\end{algorithm}

\newpage

\textcolor{black}{\subsection{Batch Bayesian Optimization Algorithm (q-BO)}
The q-BO or batch Bayesian optimization algorithm uses a multipoint acquisition function, $AF^{\ell}_q(x_{\mathcal{K}}; \kappa)$, like the q-LCB presented in \cite{wilson2017qBO}, to select a batch of $q$ experiments that can be run in parallel. Unlike most adaptations of BO where the AF is optimized over a single point, the q-LCB is optimized over a set of $q$ points. By selecting the experiments in a batch rather than independently, as in HP-BO, or sequentially, as in MC-BO, q-LCB is able to measure the correlation between the suggested sample locations, allowing it to more easily avoid the issue of redundant sampling. Given a desired batch size, the value of a particular set of experiments $x_{\mathcal{K}}$ is measured according to:
\begin{equation}\label{eq:qLCB}
	AF^{\ell}_q\left(x_{\mathcal{K}}; \kappa\right)=\frac{1}{S}\sum_{s=1}^{S}\textrm{max}\left(\mu^{\ell}_q\left(x_{\mathcal{K}}\right)-\kappa\cdot|A^{\ell}_q\left(x_{\mathcal{K}}\right)z_s|\right)
\end{equation}
where $\mu^{\ell}_q\left(x_{\mathcal{K}}\right)\in\mathbb{R}^q$ and $A^{\ell}_q\left(x_{\mathcal{K}}\right)\in\mathbb{R}^{q\times q}$ are the GP mean and Cholesky factor of the GP covariance ($AA^{T}=\Sigma$) at a batch of points $x_{\mathcal{K}}$ respectively, $z_s\in\mathbb{R}^q$ is a random variable with $z_s\sim\mathcal{N}(\boldsymbol{0}, \boldsymbol{I})$, and $|\cdot|$ is the absolute value (element-wise) operator. The new batch is then selected by solving:
\begin{subequations}\label{eq:qBO}
	\begin{gather}
		x^{\ell+1}_{\mathcal{K}}\leftarrow \mathop{\textrm{argmin}}_{x_{\mathcal{K}}}~~AF^{\ell}_q\left(x_{\mathcal{K}}; \kappa\right)\\
		\textrm{s.t.}~~x_{\mathcal{K}}\in{X}
	\end{gather}
\end{subequations}
where, again, the optimization is done over the entire batch of $q$ points in $x_{\mathcal{K}}$. The experiments are then run (in parallel) and the collected performance measurements are used to update the dataset $\mathcal{D}^{\ell+1}\leftarrow \mathcal{D}^{\ell}\cup \{x^{\ell+1}_{\mathcal{K}}, f^{\ell+1}_{\mathcal{K}}\}$. This data is used to retrain the model which enables the selection of the next batch of experiments. The pseudocode for q-BO is summarized in Algorithm \ref{alg:q-BO}.}
\\

\textcolor{black}{The q-BO algorithm has proven to be especially popular in the multi-objective optimization setting and is, in principle, not difficult to implement. However, unlike single-point AFs, multi-point AFs do not have a closed-form representation and, as a result, constructing and optimizing $AF^{\ell}_q(x_{\mathcal{K}}; \kappa)$ requires the use of numerical methods like Monte Carlo, as seen in \eqref{eq:qLCB}, making this an intensive process, especially as $q$ increases. Additionally, while the use of the Cholesky factor ensures that the algorithm cannot select redundant experiments, safeguards must be placed when constructing the AF optimization problem to ensure that the optimizer cannot select identical points as this will result in the covariance matrix $\Sigma\left(x_{\mathcal{K}}\right)$ being singular and cause the optimizer to fail. In this work, that safeguard was implemented as a tolerance value $\epsilon$ that set the minimum allowable distance between any two points within $x_{\mathcal{K}}$. We should note that while using this strategy we observed that q-BO can and does occasionally select points that are within $\epsilon$ of each other. This can be practically as undesirable as redundant sampling, depending on the value of $\epsilon$. A reference model can also be easily incorporated into the q-BO approach by having the algorithm learn the residual instead of the performance function.}

\RestyleAlgo{ruled}
\begin{algorithm}[!hbt]
\caption{Batch Bayesian Optimization Algorithm (q-BO)}\label{alg:q-BO}
Given $\kappa$, $S$, $L$, and $\mathcal{D}^{\ell}$\;
Train GP $\hat{f}^\ell$ using initial dataset $\mathcal{D}^{\ell}$\;
\For{$\ell=1, 2,..., L$}{
Compute $x_{\mathcal{K}}^{\ell+1} \gets \mathop{\textrm{argmin}}_{x_{\mathcal{K}}}{AF_q^{\ell}(x_{\mathcal{K}};\kappa)}$ s.t. $x_{\mathcal{K}}\in X$\;
\For{$k\in \mathcal{K}$}{
Evaluate performance at $x_{k}^{\ell+1}$ to obtain $f^{\ell+1}_{k}$\;
}
Update data $\mathcal{D}^{\ell+1}\gets \mathcal{D}^{\ell}\cup  \left\{x_\mathcal{K}^{\ell+1}, f_\mathcal{K}^{\ell+1}\right\}$\;
Train GP using $\mathcal{D}^{\ell+1}$ to obtain $\hat{f}^{\ell+1}$\;
}
\end{algorithm} 

%%%%%%%%%%%%%%%%%%%%%%%%%%%%%%%%%%%%%%%%%%
%\newpage 

\section{Parallel Bayesian Optimization using Informed Partitioning}

We propose new paradigms for parallel BO that conduct informed partitioning of the design space. Specifically, we propose a domain partitioning approach (analogous to HS-BO) that conducts partitioning by following the level sets of the performance function. Because the performance function cannot be easily evaluated, we use a reference model to guide the partitioning; this approach allows us to leverage expert or physical knowledge, which might highlight certain regions of the design space that are promising or non-promising (and with this prioritize). We also propose a variable partitioning approach that aims to exploit partially separable structures that are commonly found in complex systems; specifically, in these systems, the performance function is composed of a collection of functions for different subsystems (but the functions are coupled together via common variables).  The key idea is then to search the design space by following this separable structure, while sharing information between the coupling variables. We refer to these paradigms as level-set partitioning (LS-BO) and variable partitioning (VP-BO). 

\subsection{Level-Set Partitioning Algorithm (LS-BO)}

\begin{figure}[!htp]
	\begin{center}
		\includegraphics[height=1.85in]{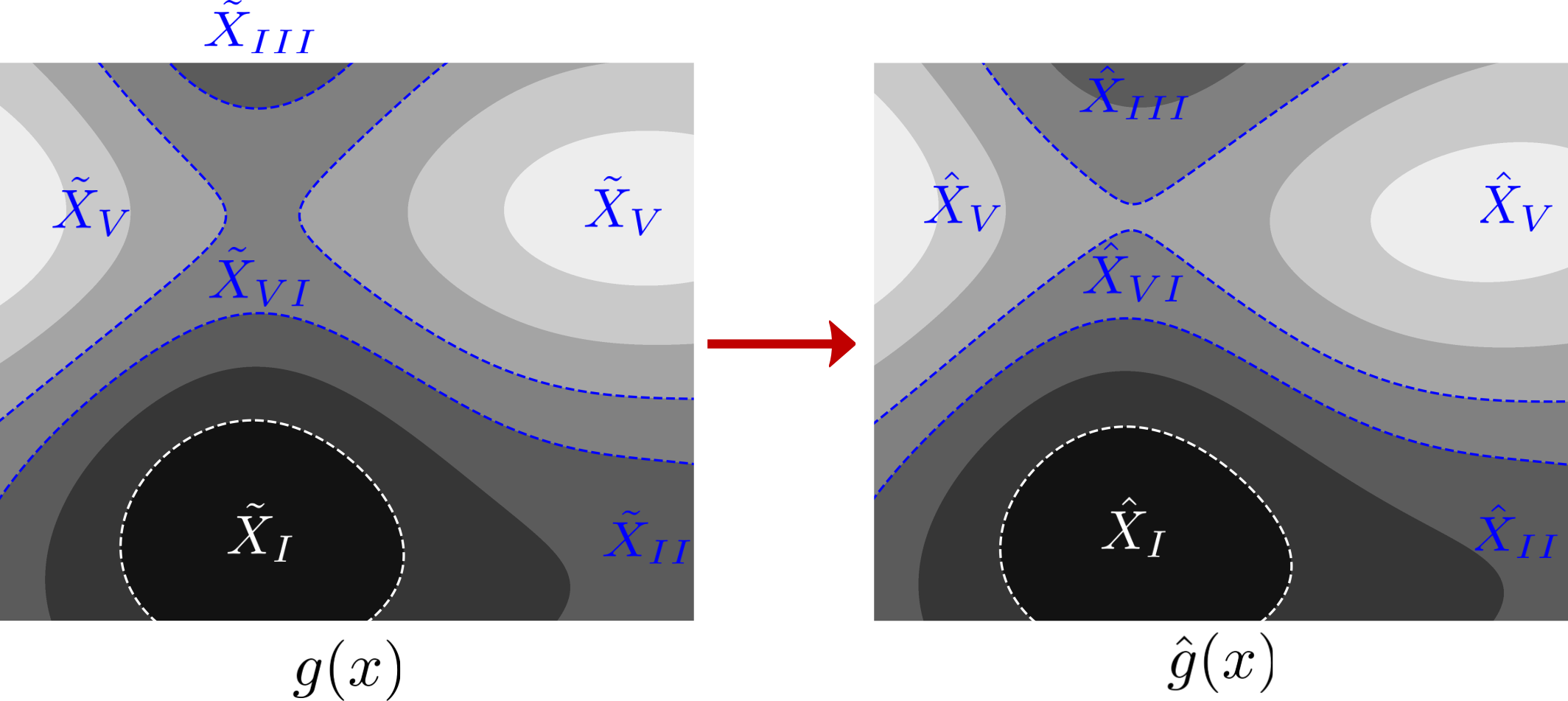}
		\caption{Level set partitioning (LS-BO) uses the $\alpha$-level sets of the reference $g$ to split $X$ into subdomains $\tilde{X}_k,\ k\in\mathcal{K}$. Depending on the complexity of $g$, enforcing level set constraints in the AF optimization problem can be difficult; therefore, the level sets are approximated using the surrogate model $\hat{g}$.}
		\label{fig:LS_BO}
	\end{center}
\end{figure}

LS-BO uses domain partitions of the design space $X$ that follow the levels sets of the reference function $g$. We recall that the $\alpha$-level set (sublevel) of this scalar function is:
\begin{align}
\textcolor{black}{\tilde{X}(\alpha)}:=\{x\in X\,|\, g(x)\leq \alpha\}\subseteq X
\end{align}
for any $\alpha\in \mathbb{R}$. We now note that solving the AF optimization problem: 
\begin{subequations}\label{part_LCB_test}
    \begin{gather}
        \min_{x}~~AF(x)\\
        \textrm{s.t.}~~ x\in \textcolor{black}{\tilde{X}(\alpha)}
    \end{gather}
\end{subequations}
would force the BO algorithm to restrict the search over a restricted subdomain $\textcolor{black}{\tilde{X}(\alpha)}$. However, solving this optimization problem can be difficult if $g$ does not have an explicit algebraic form (e.g., low-fidelity simulator) or has a complex form (e.g., physics model). To overcome this limitation, we construct a GP model $\hat{g}$ of $g$ to define the approximate level set:
\begin{align}
\hat{X}(\alpha):=\{x\in X| \hat{g}(x)\leq \alpha\}.
\end{align}
We use the previous basic observations to derive our domain partitioning approach; we construct a set of domain partitions $\hat{X}_k\subseteq X,\ k\in \mathcal{K}$ by following different level-sets of the function. Specifically, we construct the subdomains:
\begin{align}
\hat{X}_k:=\{x\in X| \alpha_{k}\leq \hat{g}(x)\leq \alpha_{k+1}\},\; k\in \mathcal{K}, 
\end{align}
We note that the subdomains are upper and lower bounded in order to obtain non-overlapping partitions. \textcolor{black}{The level set thresholds $\alpha_k$ are set by discretizing the range of $\hat{g}(x)$. The simplest method for generating the subdomains is to uniformly discretize the interval between the extreme lower and upper values of the reference model as follows:
\begin{align}
\alpha_k = \alpha_1+(k-1)\Delta,~ k=2,...,K
\end{align}
where $\Delta=\frac{\alpha_{K+1}-\alpha_1}{K}$, $\alpha_1=\mathop{\textrm{min}}_{x\in X}\hat{g}(x)$, and $\alpha_{K+1}=\mathop{\textrm{max}}_{x\in X}\hat{g}(x)$. In cases where additional specificity is desired, the partitions can be further adapted by setting the intervals according to various factors such as a focus on a particular region of the design space, the desired level of exploration vs exploitation, the level of confidence in the quality of the reference model, the geometry of $\hat{g}(x)$, and so on.} The partitioning approach is illustrated in Figure \ref{fig:LS_BO}. 

As in S-BO, we begin with dataset $\mathcal{D}^\ell=\{x_\mathcal{K},f_\mathcal{K}\}$ and build a GP $\hat{f}^\ell$ and the acquisition function $AF^\ell$. We then obtain a new set  $x_\mathcal{K}^{\ell+1}$ by solving the following collection of optimization problems:
\begin{subequations}\label{part_LCB2}
    \begin{gather}
        x_k^{\ell+1}\leftarrow \mathop{\textrm{argmin}}_{x}~~AF^\ell(x)\\
        \textrm{s.t.} ~~x\in \hat{X}_k
    \end{gather}
\end{subequations}
for $k\in \mathcal{K}$.  Using the new experiments $x_\mathcal{K}^{\ell+1}$ we evaluate system performance $f_\mathcal{K}^{\ell+1}$ (in parallel) and we append the collected data to the dataset $\mathcal{D}^{\ell+1}\leftarrow \mathcal{D}^\ell \cup \{x_\mathcal{K}^{\ell+1},f_\mathcal{K}^{\ell+1}\}$. The new dataset is used to update the GP $\hat{f}^{\ell+1}$ and the acquisition function $AF^{\ell+1}$. \textcolor{black}{A summary of this procedure is presented in Algorithm \ref{alg:LS-BO}.}

\begin{algorithm}[!htp]
\caption{Level-Set Partitioning BO (LS-BO)}\label{alg:LS-BO}
Given $\kappa$, $g$, $L$, $K$\;
Build surrogate $\hat{g}$ of $g$\;
Construct partitions $\hat{X}_k\subseteq X,\, k\in \mathcal{K}$ using level sets of $\hat{g}$\;
Train GP $\hat{f}$ with initial dataset $\mathcal{D}^{\ell}$\;
\For{$\ell=1, 2,..., L$}{
\For{$k\in \mathcal{K}$}{
Compute experiment $x_k^{\ell+1}\gets\mathop{\textrm{argmin}}_{x}{AF^{\ell}_{f}(x;\kappa)}$ s.t. $x\in \hat{X}_k$\;
Evaluate performance at $x_k^{\ell+1}$ to obtain $f_k^{\ell+1}$\;} 
Update $\mathcal{D}^{\ell+1}\gets \mathcal{D}^{\ell}\cup  \left\{x_\mathcal{K}^{\ell+1}, f_\mathcal{K}^{\ell+1}\right\}$\;
Retrain GP using $\mathcal{D}^{\ell+1}$ to obtain $\hat{f}^{\ell+1}$\;}
\end{algorithm}

It is important to highlight that the LS-BO approach that we propose uses a GP model of the performance function and an AF that are defined over the entire design space $X$; this approach thus differs from HS-BO (which uses a different GP and AF in each partition $X_k$). Moreover, we note that the partitioning of the space follows the level sets of the reference function, and this allows us to concentrate experiments over regions that are most promising. The proposed LS-BO approach can also be implemented in such a way that the reference function is exploited to learn the residual (as opposed to learning the performance function). As such, we can leverage the reference function for constructing the domain partitions and for guiding the search. 

\subsection{Variable Partitioning Algorithm (VP-BO)}

Many physical systems are typically composed of individual components that are partially interconnected (e.g., they are modular). For instance, a chemical process includes units (e.g., reactors and separations) that are interconnected, and the performance of each unit contributes to the total system performance. Moreover, the performance of each unit is typically strongly affected by the unit variables and less affected by variables of other units. This partially separable structure can be captured as the following optimization problem:
\begin{subequations}\label{separable}
    \begin{gather}
        \min_{x}~~\sum_{k\in\mathcal{K}}f_k(x_k;\mathbf{x}_{-k})\\
        \textrm{s.t.}~~ x\in X
    \end{gather}
\end{subequations}
where $f_k:X\to \mathbb{R}$ is the performance contribution of component $k$, the entire set of decision variables is split into $K$ subsets as $x:=\{x_1,x_2,....,x_{K}\}$, and we define $\mathbf{x}_{-k}:=x\setminus \{x_k\}$ (entire set of variables that does not include $x_k$). \textcolor{black}{We should note that the variable partitions should be non-overlapping subsets (i.e., $x_i \cap x_{j\neq i} = \{0\}, i, j \in \mathcal{K}$).} Additionally, we assume that that performance function can be decomposed as $f=\displaystyle \sum_{k\in\mathcal{K}}f_k$. 
\\

VP-BO follows a Gauss-Seidel paradigm; assume we have an initial set of data $\mathcal{D}^\ell=\{x_{\mathcal{K}}^\ell, f_{\mathcal{K}}^\ell\}$; Here, \textcolor{black}{we assume that we measure $f_1,...,f_K$ in each experimental module so that $f_\mathcal{K}^{\ell} \in \mathbb{R}^{\ell \times K}$ rather than $\mathbb{R}^{\ell}$ as in the previous algorithms; note that this means that $f_k^\ell$ corresponds to the $k^{th}$ column of $f_\mathcal{K}^{\ell}$.} We optimize the individual performance of subcomponent $k$ using the variables $x_k$, while keeping the rest of the variables $\mathbf{x}_{-k}$ constant (to the values of the previous iteration $\ell$):
\begin{subequations}\label{part_var_goal}
    \begin{gather}
        \min_{x_k}~~f_k(x_k;\mathbf{x}_{-k}^\ell)\\
        \textrm{s.t.}~~(x_k;\mathbf{x}_{-k}^\ell) \in{X}
    \end{gather}
\end{subequations}
for $k\in \mathcal{K}$. Accordingly, we decompose the AF optimization problem into the subproblems:
\begin{subequations}\label{part_var_AF}
    \begin{gather}
        x_k^{\ell+1}\leftarrow \mathop{\textrm{argmin}}_{x_k}~~AF_{f_k}^{\ell}(x_k;\mathbf{x}_{-k}^\ell,\kappa)\\
        \textrm{s.t.}~~(x_k;\mathbf{x}_{-k}^\ell) \in{X}.
    \end{gather}
\end{subequations}
for $k\in \mathcal{K}$. Here, $AF_{f_k}^{\ell}$ is the acquisition function of component $k$ which is built using the GP $\hat{f}_k$ of the performance $f_k$.  Moreover, $\mathbf{x}_{-k}^\ell$ is the value of the variables not in partition $k$ at the current iteration (which are held fixed when optimizing the $AF_{f_k}^{\ell}$).
\\

We partition the variables by leveraging the reference model $g$; specifically, we use this the reference to identify which variables have the most impact on individual components of the system; \textcolor{black}{this can be done in various ways. The most straightforward method would be via inspection using a combination of information provided by the reference model and any available expert knowledge over the importance of the various inputs on the subsystems. If such information is not available, $g(x)$ can instead be analyzed with an appropriate feature importance technique such as sparse principal components analysis (SPCA) \cite{spca2006zou}, automatic relevance determination (ARD) \cite{williams1996gpml}, model class reliance (MCR) \cite{fisher2019mcr}, etc., to determine the appropriate variable-subsystem pairings. Because the partitions must not overlap, the results of this analysis should be checked for instances where an input is paired with multiple subsystems. If this occurs, we recommend that the input in question be paired with the subsystem where it has the highest relative importance. The pseudocode for implementing VP-BO is shown in Algorithm \ref{alg:VP-BO}.}
\\

One of the advantages of the VP-BO approach is that the AF optimization over each partition only uses a subset of variables; this can significantly reduce the computational time. Moreover, this approach is amenable for implementation in a distributed manner (e.g., each unit of a process has its own separate BO algorithm). The VP-BO approach (and the LS-BO approach) also takes system-specific behavior into account when developing partitions (informed by the reference model). As we will show in the next section, the use of prior knowledge can lead to significant reductions in computational time and in the number of experiments performed. Moreover, we will see that such knowledge can help identify solutions and surrogate models of higher quality.  VP-BO can also be implemented in such a way that reference model is also used to guide the construction of the performance function (by learning the residual instead of the performance function). We also highlight that the VP-BO approach proposed is implemented in a way that each partition has its own GP model and AF; however, it is also possible to implement this approach by building a central GP and AF that are optimized in each partition using a different set of variables. 

\begin{algorithm}[!htb]
\caption{Variable Partitioning BO (VP-BO)}\label{alg:VP-BO}
Given $\kappa$, $K$, $L$, and $\mathcal{D}^\ell$\;
Decompose $f(x)$ into $f_k(x_k, \textbf{x}_{-k})$ for $k\in \mathcal{K}$\;
Use $\mathcal{D}^{\ell}$ to train GPs $\hat{f}_k,\, k\in \mathcal{K}$\;
\For{$\ell=1, 2,..., L$}{
\For{$k\in \mathcal{K}$}{
Compute experiment $x_k^{\ell+1}\gets\mathop{\textrm{argmin}}_{x_k}{AF_{f_k}^{\ell}\left(x_k; \textbf{x}_{-k}^{\ell},\kappa\right)}$ s.t. $\left(x_k, \textbf{x}_{-k}^{\ell}\right)\in X$\;
\textcolor{black}{Evaluate $f_1,...,f_K$ at $x_k^{\ell+1}$ to obtain $f_{\mathcal{K}}^{\ell+1}[k, :]$}\;}
\For{$k\in \mathcal{K}$}{
\textcolor{black}{$\textbf{x}_{-k}^{\ell+1}\gets\mathop{\textrm{argmin}}_{\textbf{x}_{-k}}{f_\mathcal{K}^{\ell+1}[:, k]}$}\;}
Update $\mathcal{D}^{\ell+1}\gets \mathcal{D}^{\ell}\cup  \left\{x_\mathcal{K}^{\ell+1}, f_\mathcal{K}^{\ell+1}\right\}$\;
Retrain GPs $\hat{f}_k,\,k\in\mathcal{K}$ using $\mathcal{D}^{\ell+1}$
}
\end{algorithm}

%%%%%%%%%%%%%%%%%%%%%%%%%%%%%%%%%%%%%%%%%%
\section{Numerical Case Studies}

We now present numerical results using the different BO strategies discussed; our goal is to demonstrate that the parallel BO approaches proposed provide significant advantages over S-BO and over other state-of-the-art parallel approaches. Our study simulates the performance of a pair of reactors connected in series; the operating cost of this system is a complex function of the operating temperatures. The detailed physical model used to simulate the performance of the system is discussed in the Appendix. To guide our partitioning approaches, we develop a reference model that approximates the physical model.  All data and code needed to reproduce the results can be found at \url{https://github.com/zavalab/bayesianopt}.
\\

\begin{figure}[!htp]
	\begin{center}
	    \begin{subfigure}
	        \centering
	        \includegraphics[width=3.2in]{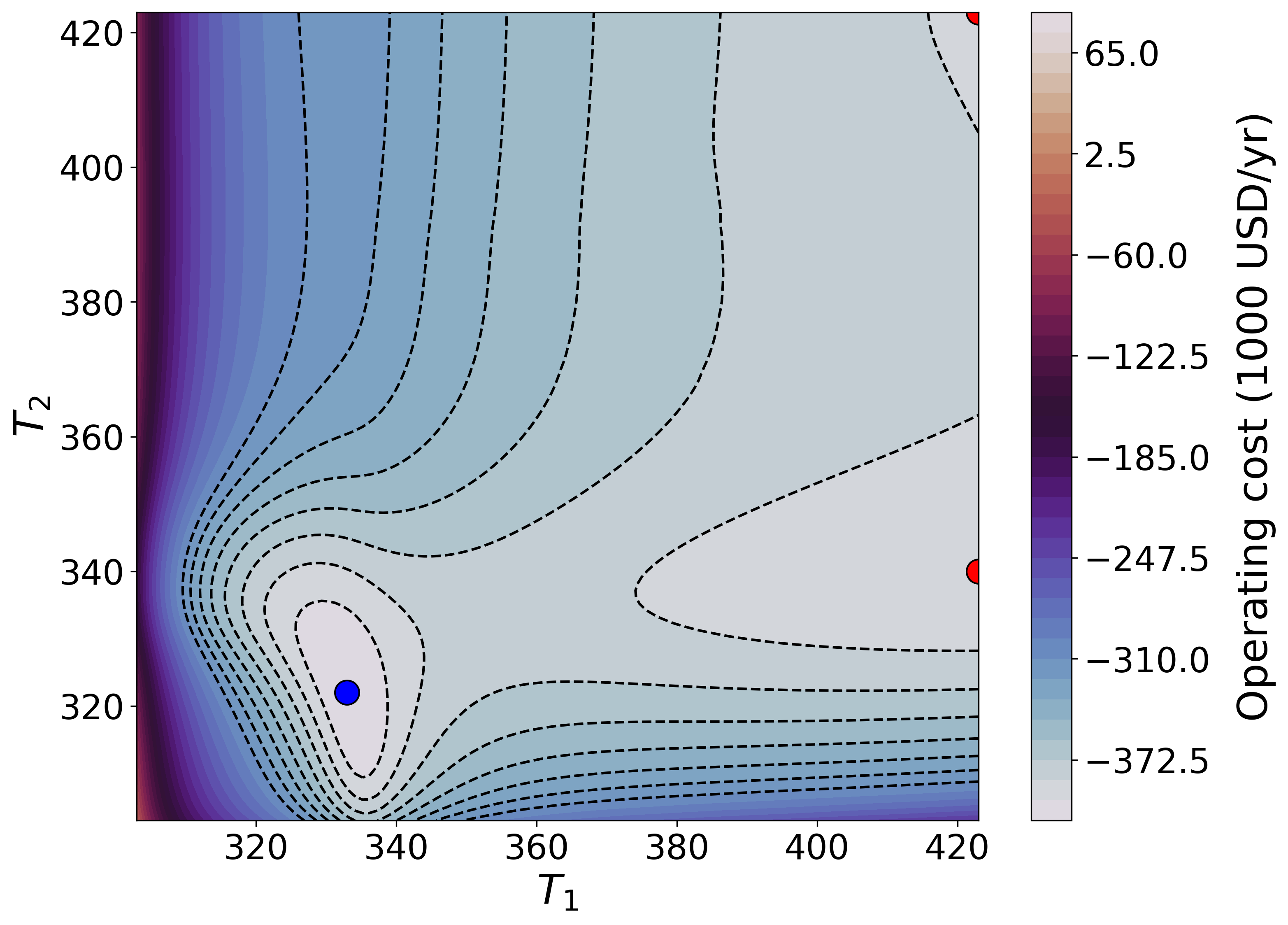}
	    \end{subfigure}
	    	    \begin{subfigure}
	        \centering
	        \includegraphics[width=3.2in]{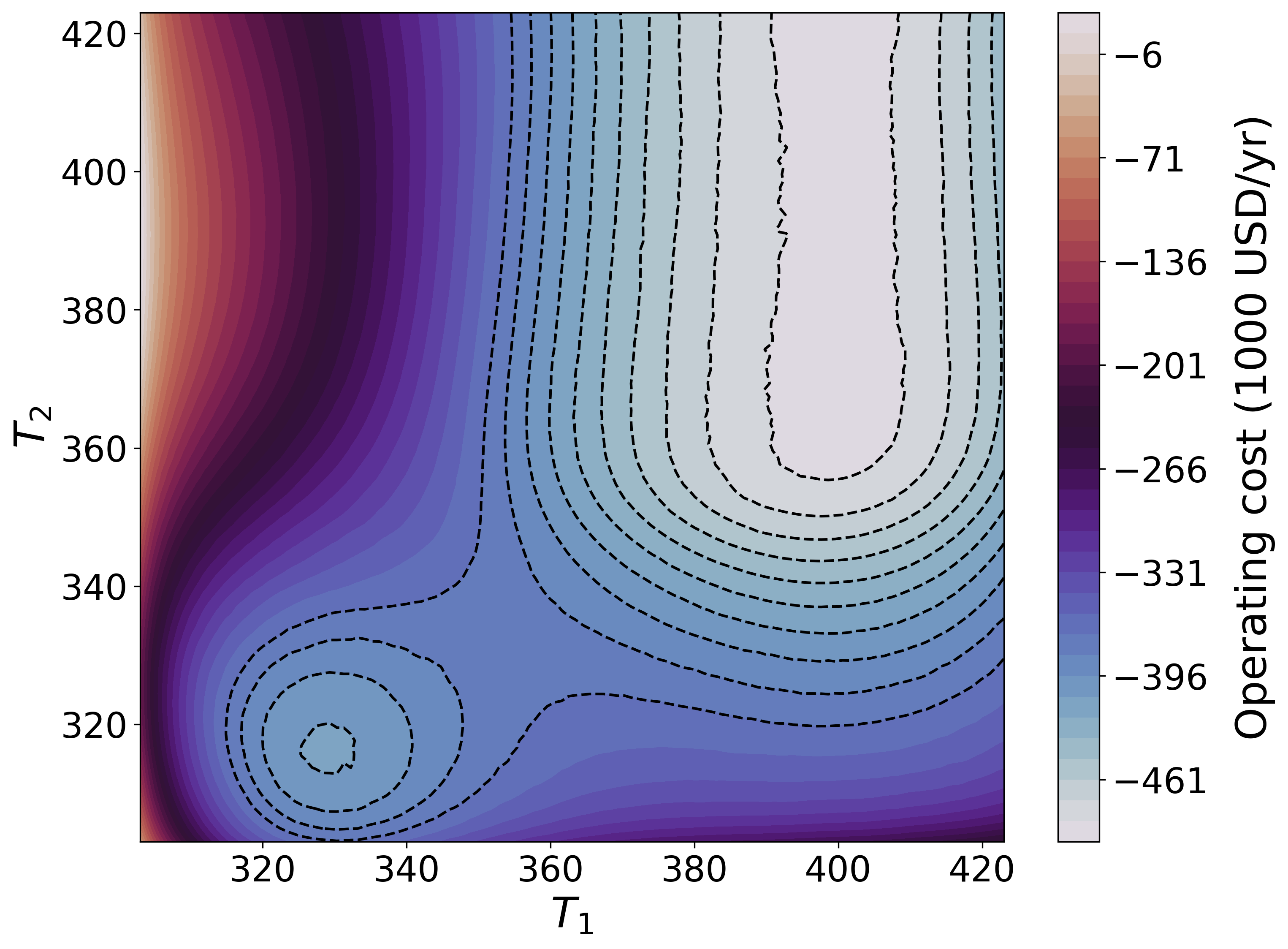}
	    \end{subfigure}
	    \vspace{-0.2in}
	    \caption{Performance function $f$ of the reactor system (left) and reference model (right). Note that the reference model captures the overall (coarse) structure of the performance function but misses some finer details.}
	    \label{fig:f_ov}
	\end{center}
\end{figure}

The optimization problem that we aim to solve with BO can be written as:
\begin{subequations}\label{rxtr_objective}
    \begin{gather}
        \min_{T_1, T_2}f(T_1, T_2)=f_1(T_1, T_2)+f_2(T_1, T_2)\\
        {\textrm s.t.}~~(T_1,T_2)\in\mathcal{T}
    \end{gather}
\end{subequations}
Figure \ref{fig:f_ov} shows the performance function $f(T_1, T_2)$ over the box domain $\mathcal{T}=\left[303, 423\right]^{2}$.  The performance function is nonconvex and contains three minima, with local solutions at $(T_1,T_2)=(423,340)$ and $(T_1,T_2)=(423,423)$ and a global solution at $(T_1, T_2)=(333,322)$. 
\\

\begin{figure}[!htp]
	\begin{center}
	    \begin{subfigure}
	        \centering
	        \includegraphics[width=3.2in]{REF_MOD.png}
	    \end{subfigure}
	    \begin{subfigure}
	        \centering
	        \includegraphics[width=3.2in]{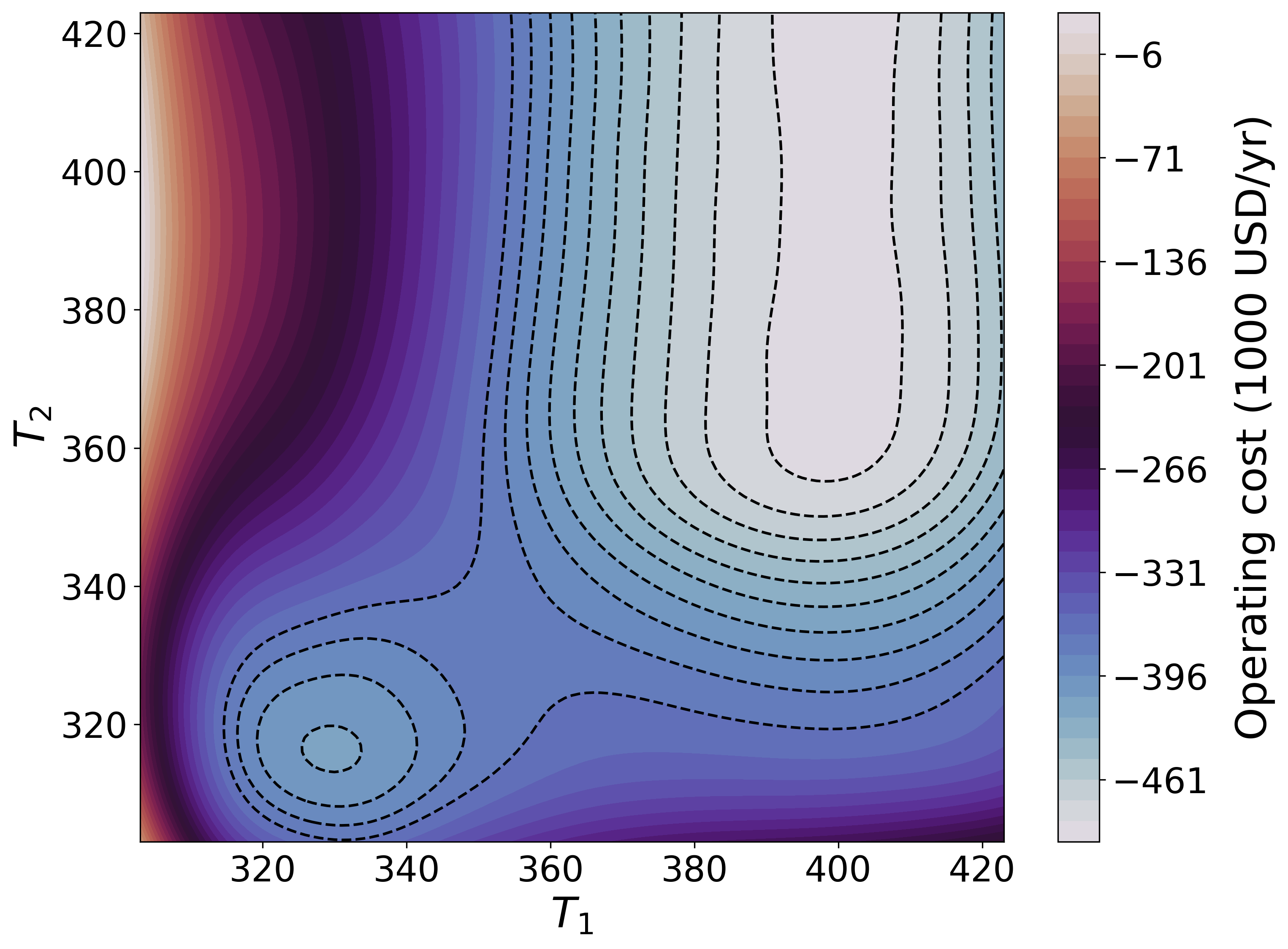}
	    \end{subfigure}
	    \vspace{-0.2in}
	    \caption{Reference model $g$ (left) and GP approximation $\hat{g}$ (right); note that the GP provides an accurate representation and can thus be used to guide partitioning approaches.}
	    \label{fig:ref_mod_and_gp}
	\end{center}
\end{figure}

The reference model $g$ is derived from a simplified physical model (see the Appendix); however, this model would be difficult to incorporate directly in the AF formulation for the LS-BO and VP-BO approaches (because the model involves a complex set of algebraic equations). As such, we approximate this model using a GP, $\hat{g}$, and use this as the reference.  Figure \ref{fig:ref_mod_and_gp} illustrates that the GP $\hat{g}$ is virtually indistinguishable from the simplified physical model $g$; as such, we can safely use this to guide our search and to guide our partitioning approaches.
\\

The HS-BO algorithm is restricted to $2^2=4$ partitions when dealing with a 2D design space; as such, and in order to achieve fair comparisons, we limit the number of parallel experiments collected with MC-BO, HP-BO, \textcolor{black}{q-BO}, and LS-BO to 4.  The VP-BO algorithm was run using 2 partitions (one for each reactor). All algorithms were implemented in Python 3.7 \textcolor{black}{and the GP modeling was done using the {\tt gaussian\_process} package in Scikitlearn. Specifically, we used the built-in {\tt Matern} method as the kernel function. This selection was motivated by the ability of the M\'atern kernel to control the smoothness of the resultant function making it highly flexible and capable of accurately modeling systems that exhibit significant nonlinearity and non-smoothness; we set the smoothness parameter $\nu=2.5$, which tends to be the standard choice. At every iteration, the optimal value of the kernel's hyperparameters, the characteristic length scales $l$, was updated using the package's built-in optimizer that sets $l$ by solving a log-marginal-likelihood (LML) problem.  A more detailed description of the {\tt gaussian\_process} package can be found in \cite{scikit-learn}.} The optimization of the AF was done in Scipy using an unconstrained minimization solver (based on L-BFGS-B) for every BO algorithm except LS-BO. The introduction of the reference GP model in the constraints of the AF minimization problem required the selection of a method capable of constrained optimization; for this, we selected SLSQP. Except for HS-BO, the exploratory parameter of the acquisition function was set to the same fixed value. All algorithms were initialized using the same starting point and we conducted 25 different runs with different starting points in order to evaluate robustness. We also ran instances of LS-BO and VP-BO with and without using a reference in the AF (for learning the residual or the performance function); this allowed us to isolate the impacts of the use of the reference model and ensure that observed performance improvements can be attributed to parallel capabilities. For both LS-BO and VP-BO, the reference function was always used to guide the selection of the partitions.
\\

\begin{figure}[!htp]
	\begin{center}
	    \begin{subfigure}
	        \centering
	        \includegraphics[width=3.2in]{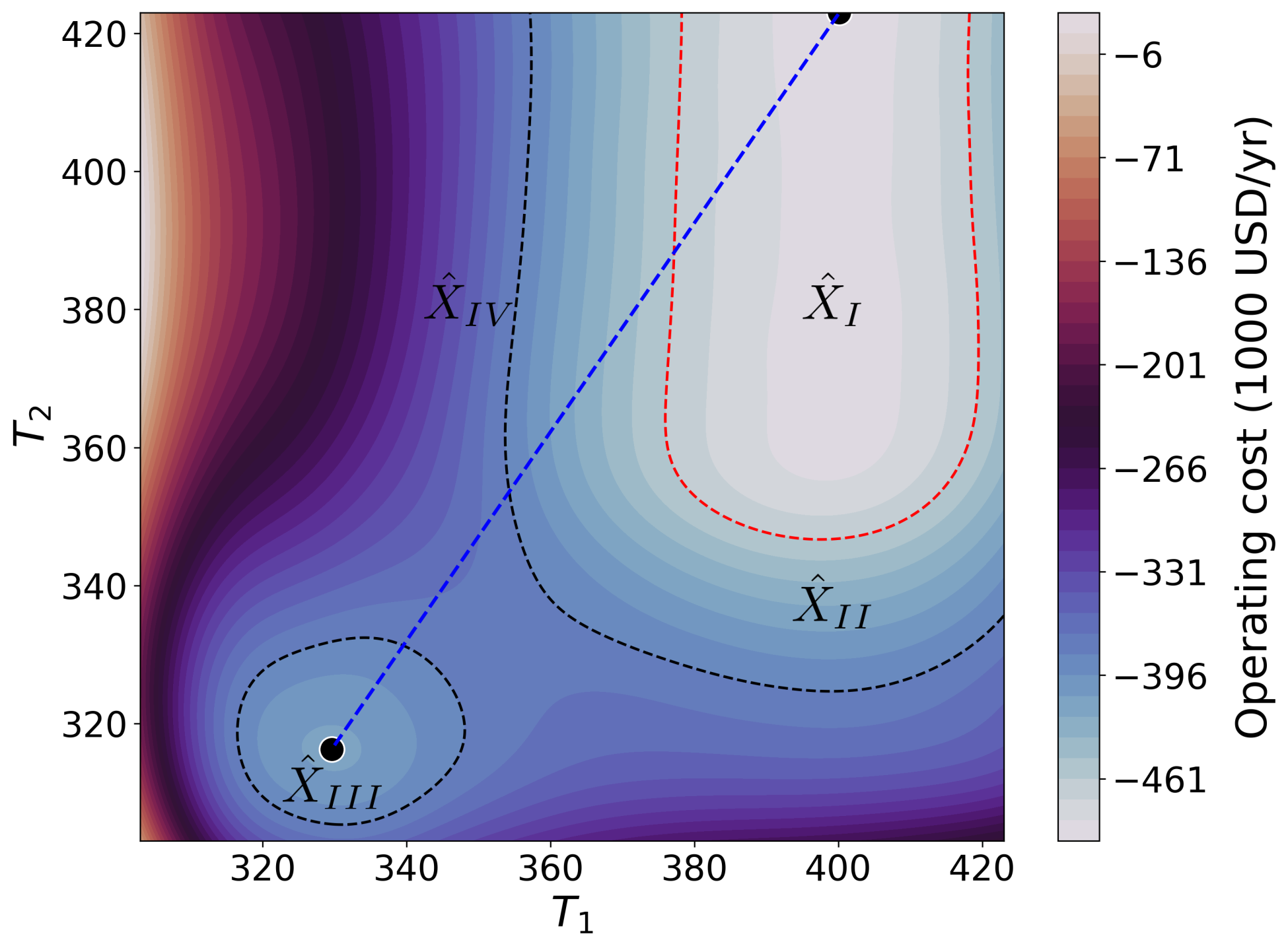}
	    \end{subfigure}
	    \begin{subfigure}
	        \centering
	        \includegraphics[width=3.2in]{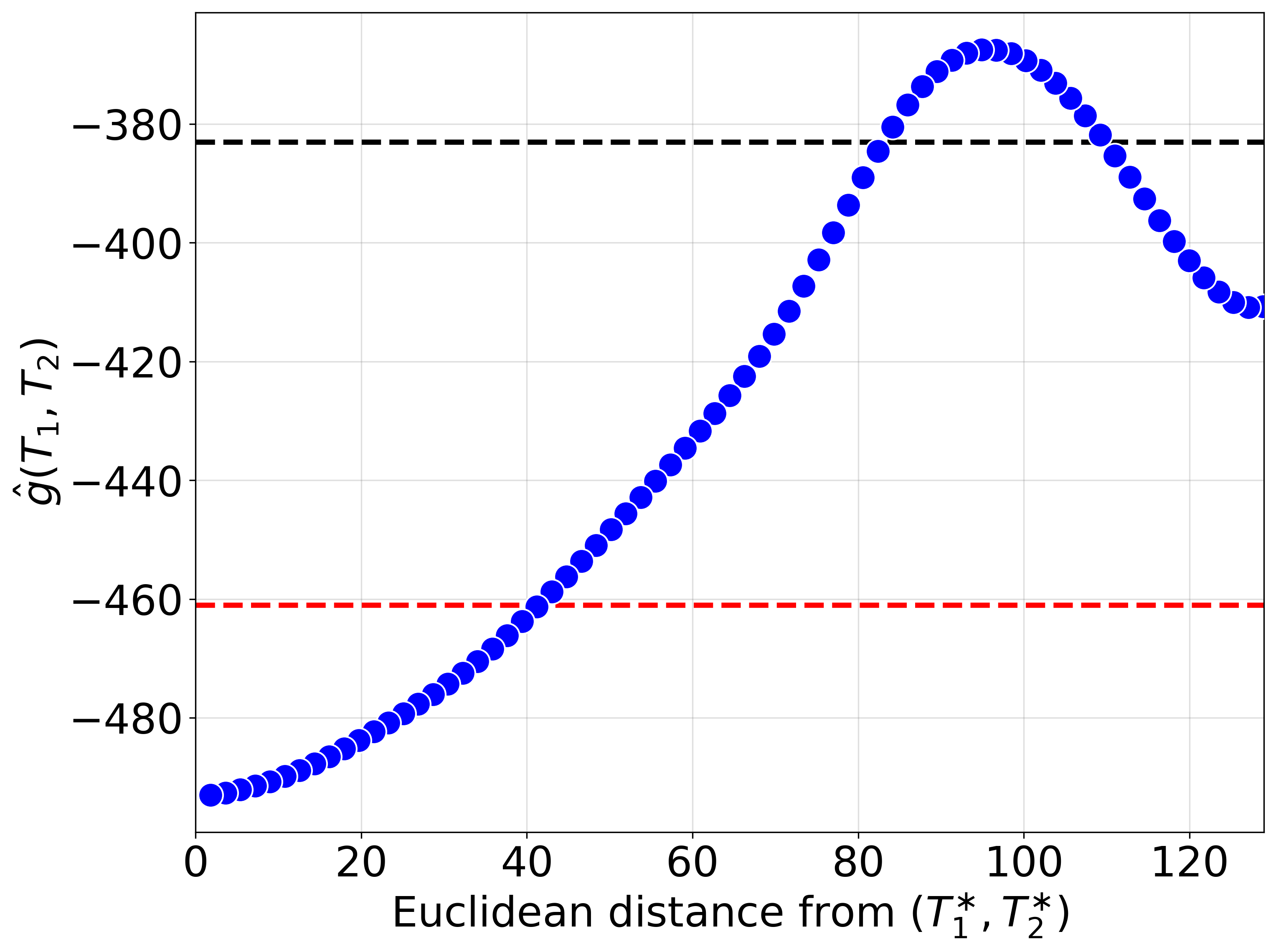}
	    \end{subfigure}
	    \caption{\textcolor{black}{Domain partitions for reactor system obtained using reference model $\hat{g}$ (left); the line connecting the two minima of the reference model is shown in blue. Values of $\hat{g}$ along this line (right) indicate that the level set $\hat{g}=-383$ (black line) provides an acceptable split between the two partitions surrounding the minima, while the level set (red line) $\hat{g}=-461$ allows for the partition surrounding the global minimum (denoted as $(T_1^{\ast}, T_2^{\ast})$) to be split into two roughly equal-sized partitions.} Note that domain $X_{III}$ is in the region of the global minimum of $f$.}
	    \label{fig:ref_mod_parts}
	\end{center}
\end{figure}

Figure \ref{fig:ref_mod_parts} highlights the level sets that we used to partition the design space for the LS-BO approach. \textcolor{black}{These partitions were generated by first locating the minima (local and global) of $\hat{g}(x)$. After determining that there were two, we discretized the range of $\hat{g}(x)$ by building a search interval around each of the minima where the lower bound of the interval was the value of the corresponding minimum. The value of $\hat{g}$ was then evaluated at various points on a line connecting the two minima to determine the spacing of the level sets. This information was used to select the upper bound of these search intervals. We were also able to use this analysis to gauge the size of both of the partitions and observed that the search region around the global minima appeared to cover a significant portion of the design space. As a result, this partition was split along the level set value that resulted in two roughly equal-sized partitions. The fourth and final partition was constructed to search the remaining space outside of the three existing partitions. Figure \ref{fig:ref_mod_parts} also provides an illustrative summary of this workflow. Note that one of the regions is near the region of the global minimum of $f$.}
\\

\begin{figure}[!htp]
	\begin{center}
	    \begin{subfigure}
	        \centering
	        \includegraphics[width=3.2in]{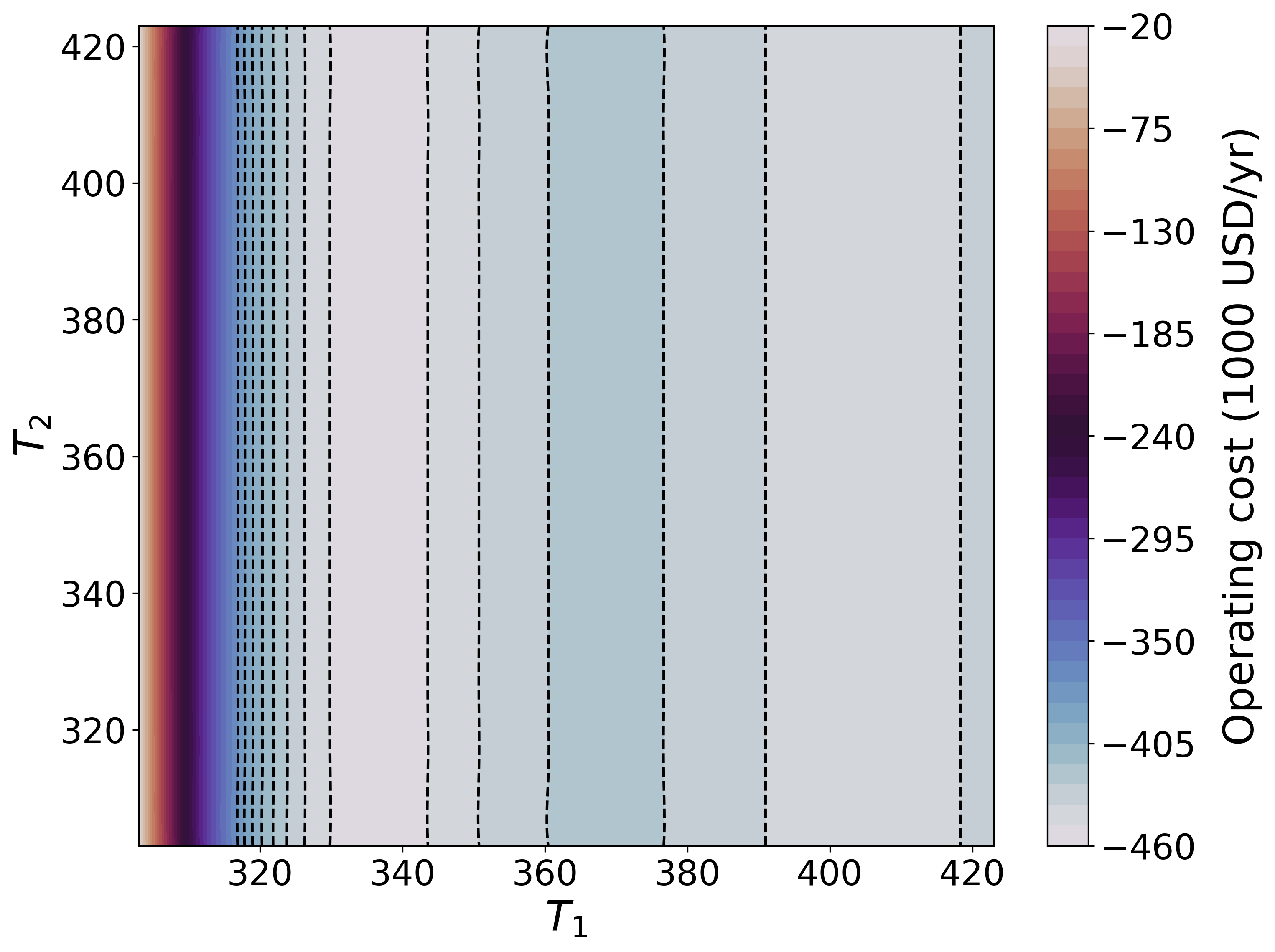}
	    \end{subfigure}
	    \begin{subfigure}
	        \centering
	        \includegraphics[width=3.2in]{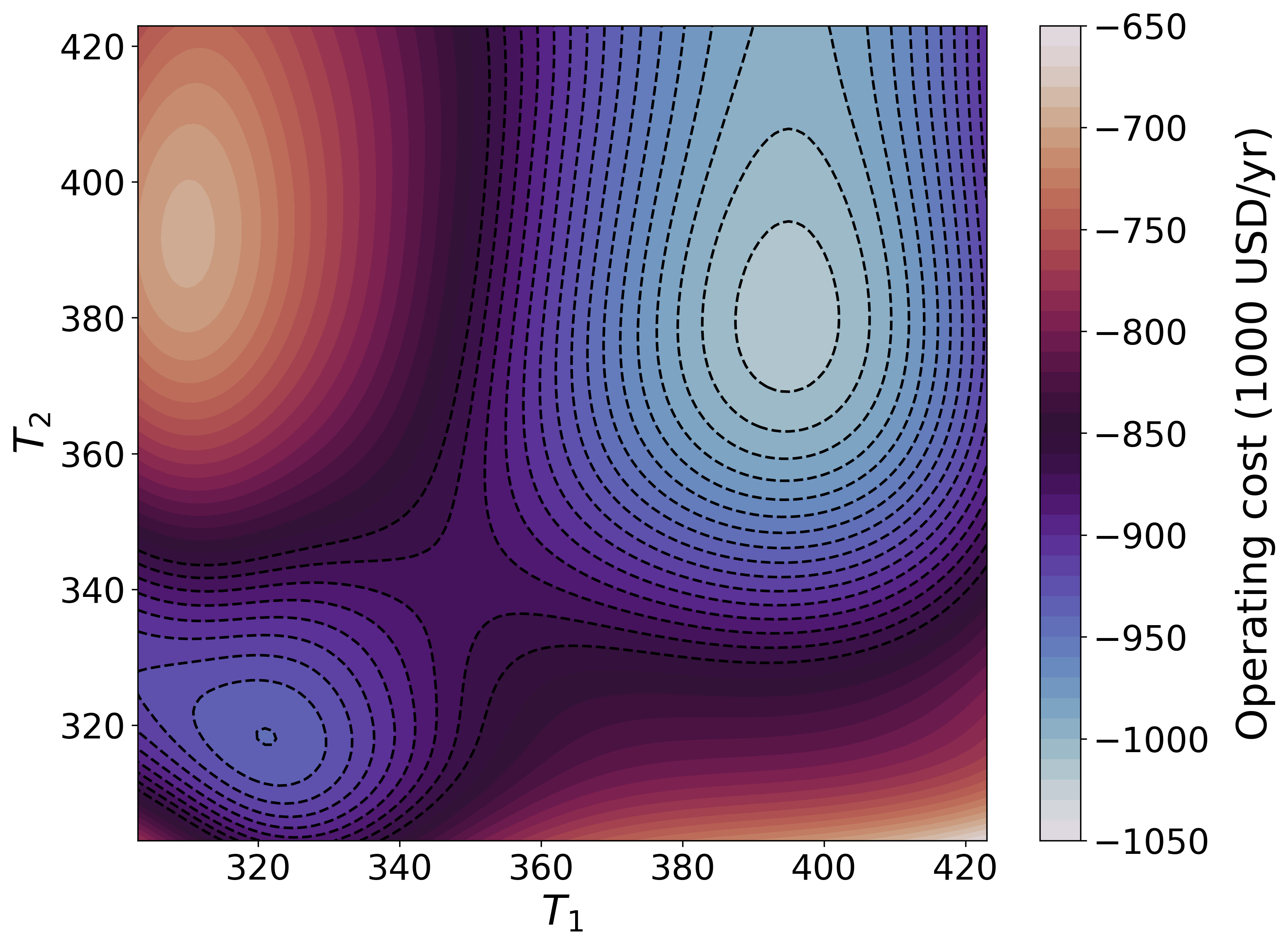}
	    \end{subfigure}
	    \caption{Reference model for the first reactor $g_1$ (left) and for the second reactor $g_2$ (right). We can see that $g_1$ is not affected by $T_2$; the combination of these functions give rise to the reference function $g=g_1+g_2$.}
	    \label{fig:g_1_and_g_2}
	\end{center}
\end{figure}

Given that the reactors are arranged in series, it is clear that the performance of the first reactor is independent of $T_2$, while the performance of the second reactor will likely have some dependence on $T_1$. Figure \ref{fig:g_1_and_g_2} demonstrates this partially-separable structure; note how the first function $g_1$ is not affected by $T_2$ (vertical lines), while $g_2$ does depend on $T_1$. \textcolor{black}{Using ARD, we confirmed that $T_1$, which had a characteristic length scale of $l=0.145$, was a more important input to $g_1$ than $T_2$ $(l=1000)$, while for $g_2$, $T_2$ $(l=0.399)$ was determined to be more important than $T_1$ $(l=0.498)$.} We thus implemented the VP-BO approach according to the following variable partitions: $x_1 = T_1$, $\bold{x}_{-1} = T_2$ and $x_2 = T_2$, $\bold{x}_{-2} = T_1$. 
\\

\begin{figure}[!htp]
	\begin{center}
	    \begin{subfigure}
	        \centering
	        \includegraphics[width=5in]{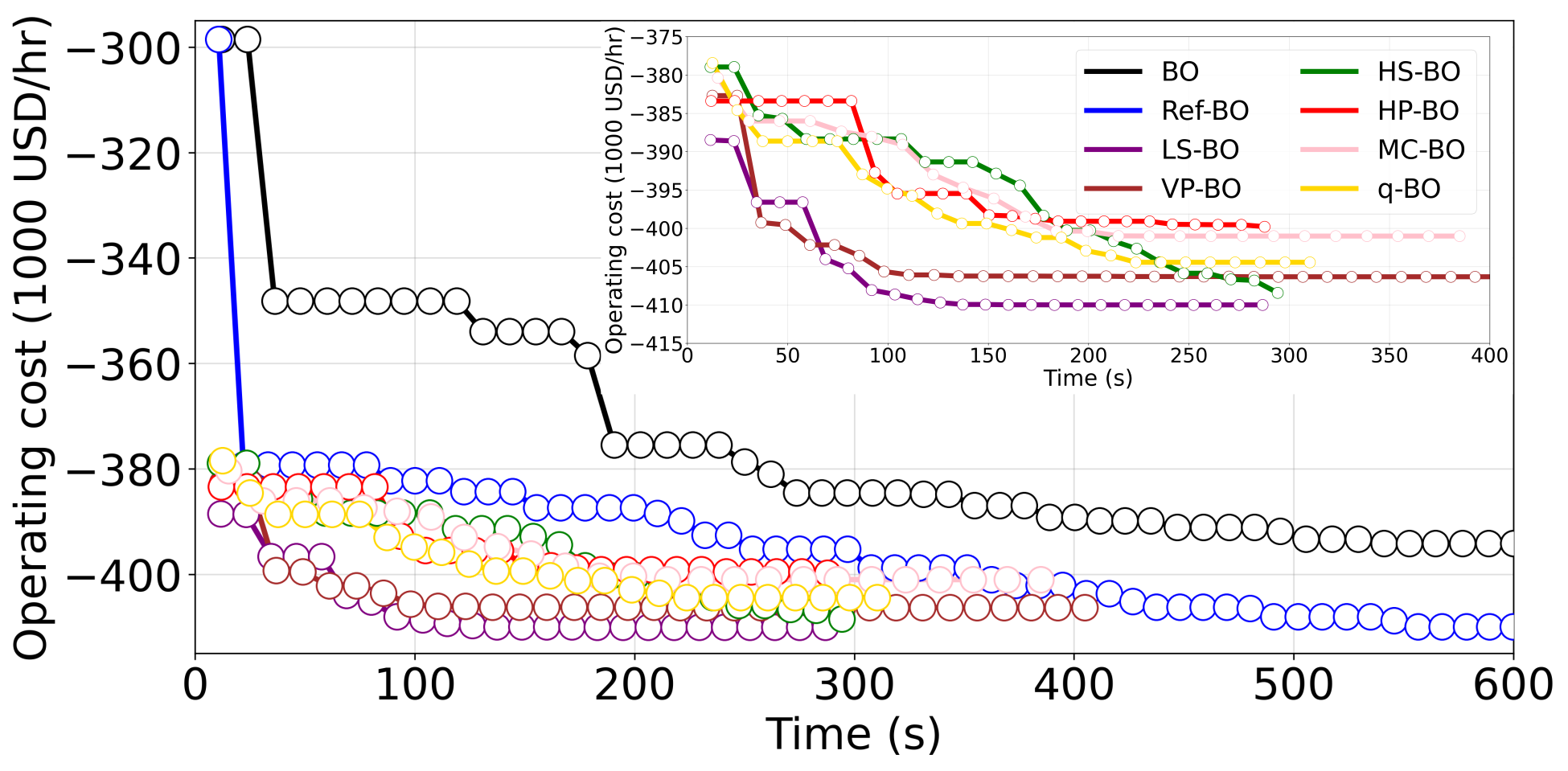}
	    \end{subfigure}
	    \caption{Total experiment time against value of best solution for tested algorithms. LS-BO and VP-BO were run using the reference model to partition the domain and guide the search.}
	    \label{fig:convergence}
	\end{center}
\end{figure}

Figure \ref{fig:convergence} summarizes the average performance (over the 25 runs) of LS-BO and VP-BO (using reference models) along with the remaining algorithms; here, we visualize the total experiment time (wall-clock time needed to evaluate performance function) against the best found performance up to the corresponding time.  Overall, we observe that all parallel BO variants performed better than standard BO both in terms of speed and best performance found. The performances at the local minima for $(T_1,T_2)=(423,340)$ and $(T_1,T_2)=(423,423)$ were approximately -395,000 USD/yr and -387,000 USD/yr respectively, while the performance of the global minimum at $(T_1,T_2)=(333,322)$ was -410,000 USD/yr. On average, the best performance obtained using BO was -394,500 USD/yr, indicating that this approach converges to a local minimum most of the time. By comparison, all the parallel BO variants found a solution that, on average, was below -400,000 USD/yr. We should also note that this improvement in performance value also comes with a significant reduction in the required wall-clock time: BO takes over 500 seconds to converge to its final solution whereas all of the parallel BO variants are able to locate a better solution in approximately 200 seconds. 
\\

The magnified profiles of Figure \ref{fig:convergence} provide a better comparison between the parallel BO variants. It is clear that LS-BO and VP-BO are significantly faster than all other variants. We also observed that LS-BO, VP-BO, HS-BO \textcolor{black}{and q-BO} consistently reached the global minimum. This illustrates how the redundant sampling seen in MC-BO and HP-BO can degrade performance. Additionally, while the performances of HS-BO \textcolor{black}{and q-BO} were similar to LS-BO and VP-BO, they required significantly more experiments to reach this performance level. From these observations we can draw a couple of conclusions: the use of a reference model for both generating system-specific partitions and simplifying the learning task delivers significant benefit; and allowing the algorithm to pool the data into a single dataset that is used to build a global surrogate model increases the predictive value of this model, resulting in faster identification of optimal regions. 
\\

\begin{figure}[!htp]
	\begin{center}
	    \begin{subfigure}
	        \centering
	        \includegraphics[width=5in]{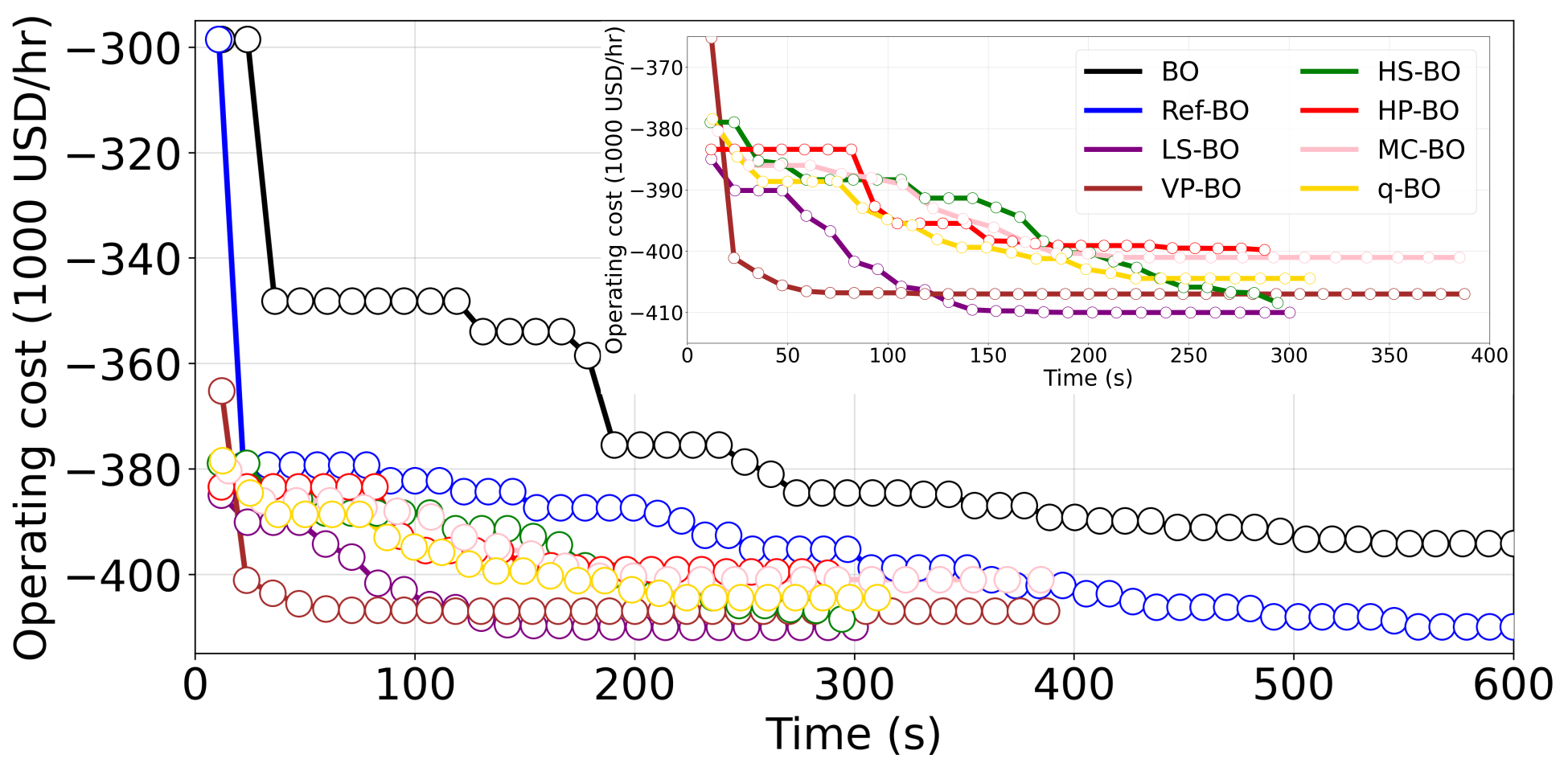}
	    \end{subfigure}
	    \caption{Total experiment time against value of best solution for the tested algorithms. LS-BO and VP-BO were run using the reference model to partition the domain but {\em not} to guide the search.}
	    \label{fig:convergence_noREF}
	\end{center}
\end{figure}

Figure \ref{fig:convergence_noREF} presents results similar to Figure \ref{fig:convergence} but we run LS-BO and VP-BO without a reference model. By comparing with the results in Figure \ref{fig:convergence} , we observe that using the reference model can help with convergence but not always. LS-BO is 24\% slower in the average convergence time, though it maintains its ability to consistently converge to the global minima. VP-BO, meanwhile, converges on average 40\% faster compared to when the reference model is used; the solution it returns is also unchanged. These results indicate that $g$ has a similar effect on LS-BO as with traditional BO, as outlined in \cite{lu2021hvac}. Namely, that it makes the search more targeted, resulting in more efficient sampling and faster conversion. Meanwhile, with VP-BO, the reference model appears to encourage more exploration of the domain, which can prevent the algorithm from converging prematurely and potentially returning a suboptimal solution. We base this claim on the fact that, when testing VP-BO without a reference, we observed that, while it is not especially sensitive to the initial values of the design variables in a given partition, it is quite sensitive to variable values of other partitions.  Overall, however, we observe that both LS-BO and VP-BO still outperform the remaining parallel algorithms (without or without a reference). These results highlight that using the proposed partitioning approaches has a larger effect on overall convergence. This allows us to confirm that the improvements we observe when using LS-BO and VP-BO can be attributed to the parallelization schemes.
\\

\begin{figure}[!htp]
	\begin{center}
	    \begin{subfigure}
	        \centering
	        \includegraphics[width=5in]{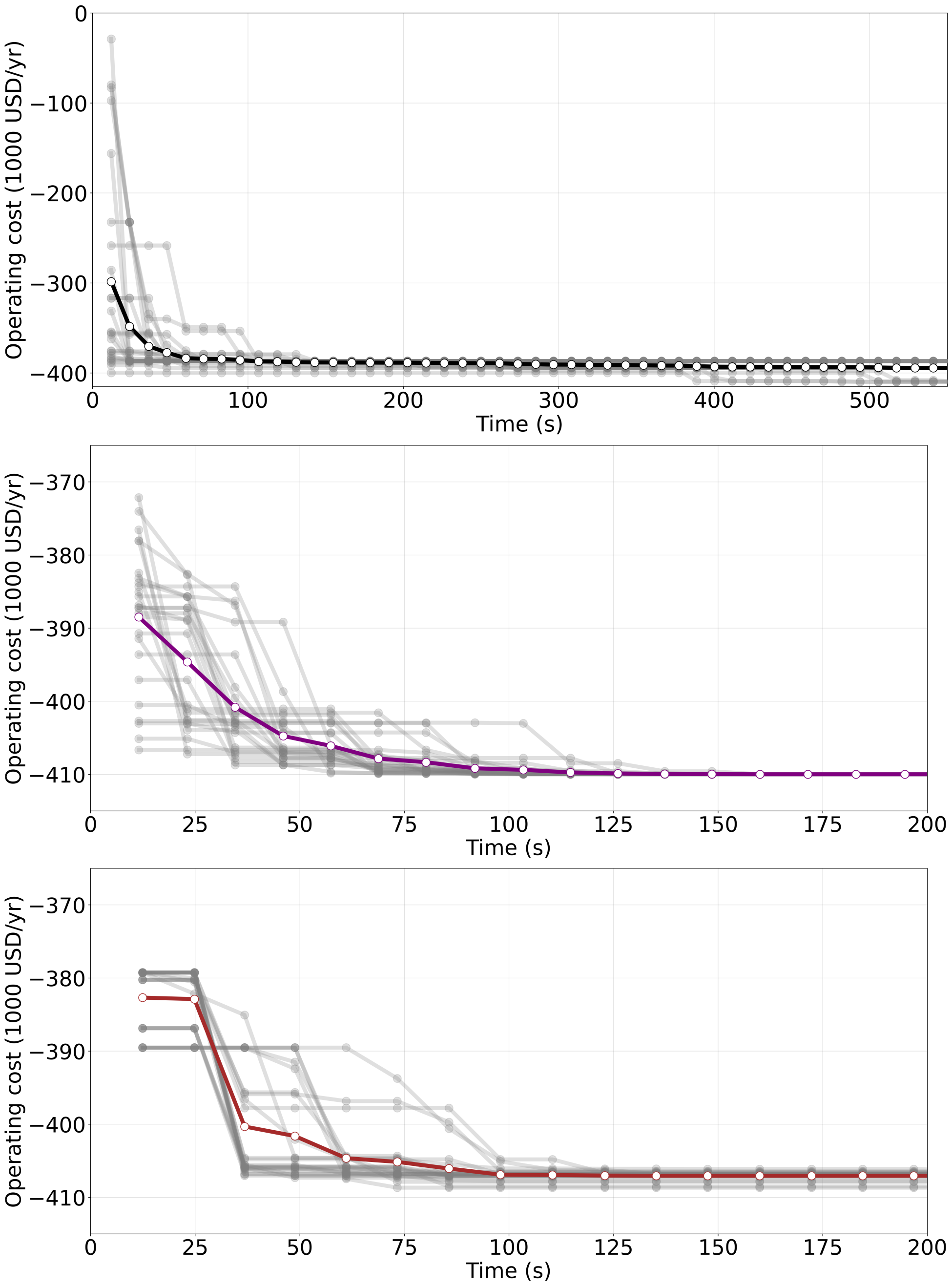}
	    \end{subfigure}
	    \caption{Distribution of the performance profiles across the 25 runs for BO (top), LS-BO (middle), and VP-BO (bottom) with the average algorithm performance is shown in color.}
	    \label{fig:convergence_dist}
	\end{center}
\end{figure}

\begin{figure}[!htp]
	\begin{center}
	    \begin{subfigure}
	        \centering
	        \includegraphics[width=5in]{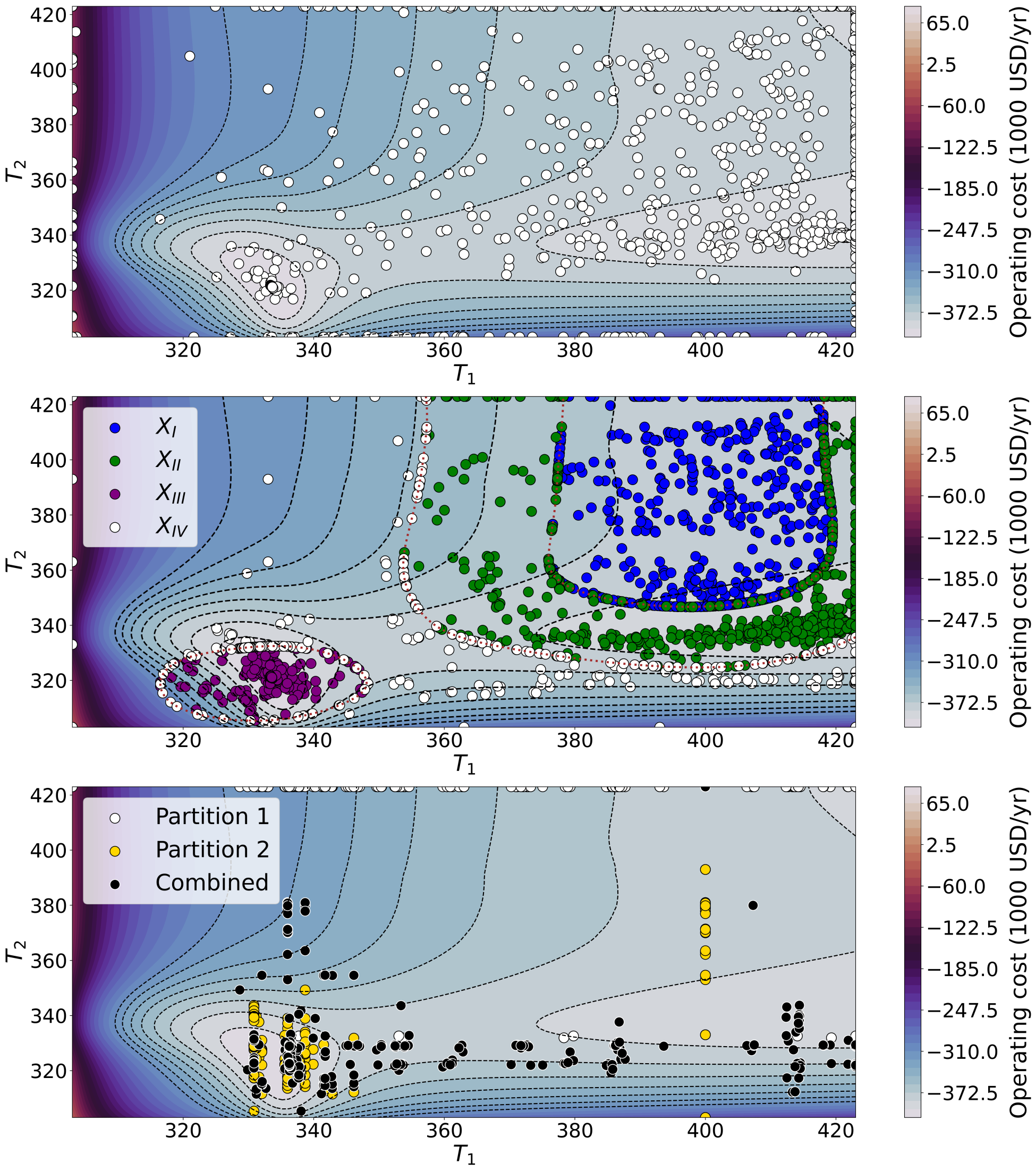}
	    \end{subfigure}
	    \caption{Experiment locations across the 25 runs for BO (top), LS-BO (middle), and VP-BO (bottom). }
	    \label{fig:sampling_dist}
	\end{center}
\end{figure}

The results presented in Figure \ref{fig:convergence} indicate that LS-BO and VP-BO are consistently more robust and sample efficient than the other approaches. The average values seen in Figures \ref{fig:convergence} and \ref{fig:convergence_noREF} provide a measure of robustness: deviations between the final reported average value and one of the three minima are due to the algorithm converging to different solutions during the various runs. For example, the final average reported value for S-BO of \$394,500 USD/yr is the result of this algorithm converging to the minima at $(T_1, T_2)=(423, 340)$ 13 out of the 25 runs, $(T_1, T_2)=(423, 423)$ for 8 runs, and to $(T_1, T_2)=(333, 322)$ the remaining 4 runs. As a result, the fact that the final reported average values for LS-BO and VP-BO are near to the global minimum indicate that these algorithms converge to or near the global solution for most if not all runs (they are robust). The convergence data collected across all runs and shown in Figure \ref{fig:convergence_dist} confirms this; we can see that, regardless of where LS-BO and VP-BO were initialized, they are always able to converge to the same region (unlike S-BO). We also see that convergence of the algorithms is in general fast but, as expected, it is sensitive to the starting point. The sensitivity to the starting point is further indication of why it is important to have expert knowledge (e.g., via use of a reference model) when initializing the search.
\\

Because evaluating the performance function tends to be expensive, reducing the number of experiments (samples) is also essential. Figure \ref{fig:sampling_dist} illustrates how S-BO, LS-BO, and VP-BO compare when it comes to sample efficiency. Standard BO samples in a significantly distributed manner with a considerable number of samples drawn from the boundaries of the domain. For LS-BO, we see that in regions where a solution exists (e.g., regions $II$ and $III$ in our case study), sampling is heavily concentrated at or near the solution. In regions where there is not a solution, the sampling is more distributed, however, the majority of samples tend to cluster around partition boundaries that are located near a solution. Samples drawn from partition $X_I$ appear to be the most widely distributed, however, this is not surprising as this partition contains a mostly flat region. Another noticeable difference when compared to traditional BO is that there is significantly less sampling at the boundaries of the domain where $f$ has unfavorable (high) values; only 8 out of 2500 samples were taken at the left and bottom bounds. VP-BO exhibits the most clustered sampling; in fact, the vast majority of the samples are drawn from or near the optimal region. Note that, while the majority of samples for $X_I$ (Partition 1) occur at the top domain boundary, this partition corresponds to reactor 1 which only depends on $T_1$ as seen in Figure \ref{fig:g_1_and_g_2}. Aside from these samples, there is a clear lack of sampling happening at the domain boundaries compared to LS-BO and traditional BO. This result, coupled with exhibiting the lowest convergence time out of all of the tested algorithms, confirm our belief that the VP-BO algorithm tends to be more exploitative. This is likely due to fact that the partitions for this algorithm are optimized over a lower dimensional space and, for a fixed $\bold{x}_{-k}$, VP-BO can find the optimal local variables $x_k$ much faster than the remaining algorithms can find an optimal global variables $x$. As a result, without the reference model to indicate the potential existence of a solution elsewhere, VP-BO seems more susceptible to settle into the first solution that it finds than algorithms like LS-BO and HS-BO whose partitions force the algorithm to search more widely. 
\\

\begin{figure}[!htp]
	\begin{center}
	    \begin{subfigure}
	        \centering
	        \includegraphics[width=5in]{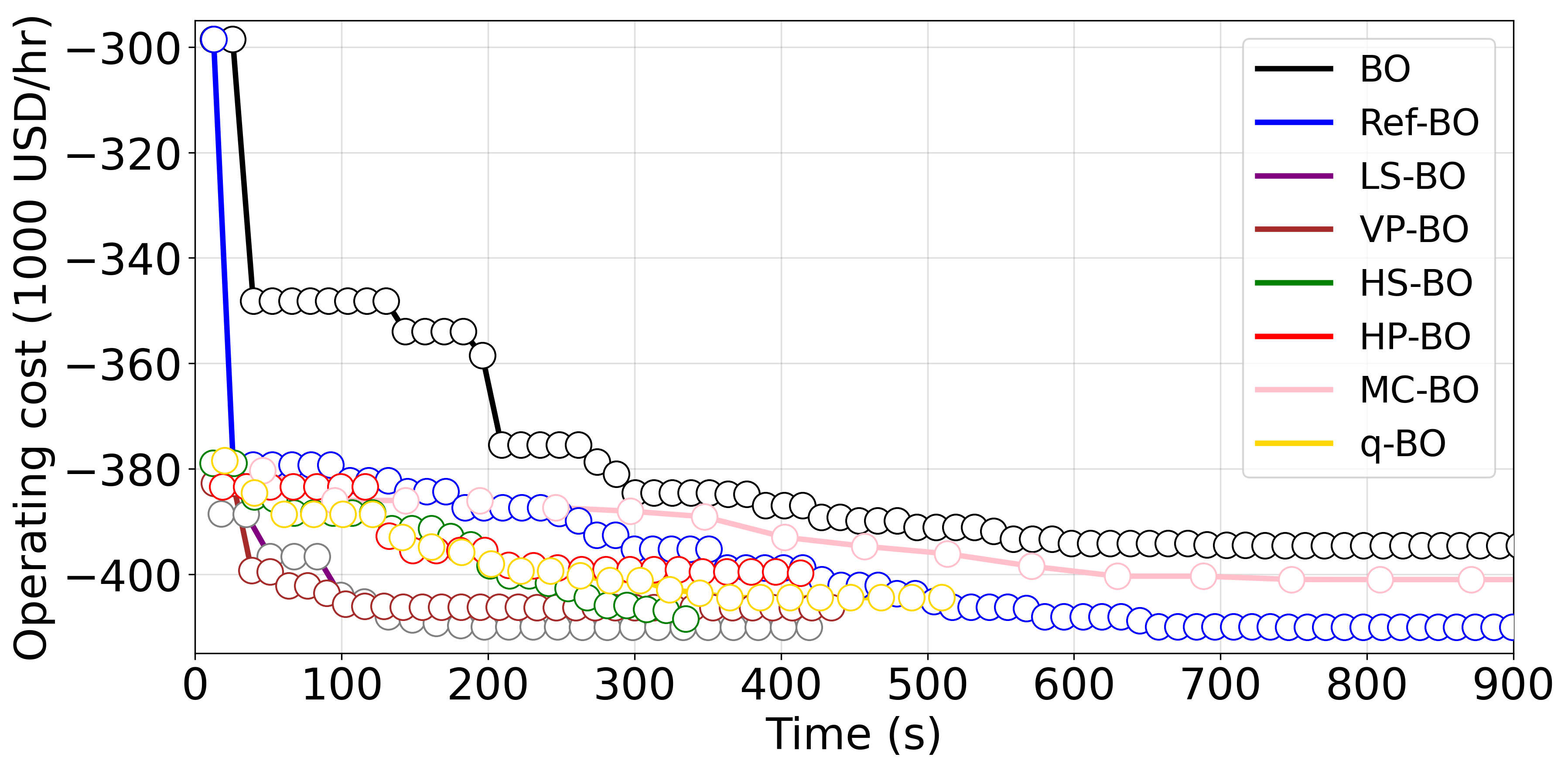}
	    \end{subfigure}
	    \caption{Profiles of wall-clock time against performance. Note that this time is comparable to the total experiment time for all algorithms; the only exception is MC-BO, indicating that the AF optimization step (and not the function evaluation) is the bottleneck for this approach.}
	    \label{fig:convergence_time}
	\end{center}
\end{figure}

To estimate the computational cost associated with the different algorithms, we measured the total wall-clock time (average across the 25 runs). The total wall-clock time includes time for performance evaluation (experiment time) and all time required to conduct other computations (e.g., AF optimization, GP training, and reference model evaluation).  The results are shown in Figure \ref{fig:convergence_time}. The closer this time is to the experimental time, the less computationally expensive the algorithm is. For instance, the total wall-clock time of S-BO was 12\% higher than the experiment time. We observe that HS-BO and VP-BO are the least computationally intensive methods, with the total wall-clock time being only 14\% and 7\% higher than the experimental time, respectively. We attribute this to the fact that HS-BO runs separate instances of BO across multiple reduced domains and, because the boundaries are rectangular, ensuring that the AF optimizer stays inside of the partition only involves bounding the upper and lower limits of $x$ it is allowed to search over, and those domains tend to be small. VP-BO, on the other hand, only optimizes over a subset of variables and this greatly reduces the time required for AF optimization. The wall-clock time of HP-BO was 44\% higher than the experiment time, this is because it requires solving multiple AF optimization problems across the entire design space. LS-BO had a total wall-clock time that was 46\% higher than the experiment time; this is attributed to the more difficult AF optimization problem that it has to solve (which has constraints defined by a GP model). \textcolor{black}{The total wall-clock time for q-BO was 64\% higher than the experiment time, which we attributed to the fact that AF optimization is done over a set of points, increasing the size of the problem that is solved. Additionally, the calculation involves more complex matrix operations and requires repetitive sampling.} MC-BO was the most computationally intensive algorithm, with a total wall-clock time that was 384\% higher than the experiment time. We attribute this to the repetitive computations in this algorithm, which require sequential sampling and GP training. 

%%%%%%%%%%%%%%%%%%%%%%%%%%%%%%%%%%%%%%%%%%

\section{Conclusions and Future Work}

We have proposed new decomposition paradigms for BO that enable the exploitation of parallel experiments. These approaches decompose the design space by following the level sets of the performance function and by exploiting the partially separable structure of the performance function. A key innovation of these approaches is the use of a reference function to guide the partitions. Using a case study for a reactor system, we have found that the proposed approaches outperform existing parallel approaches in terms of time and quality of solution found. \textcolor{black}{When using LS-BO, we observed that building partitions that are specialized beyond those that would be generated by the uniform discretization of the range of $\hat{g}(x)$, like those we used in our case study, can require significant user input. Moving forward we would like to explore methods for developing more efficient and automated protocols for generating the partitions. Additionally, we are also interested in incorporating an element of adaptivity to LS-BO and VP-BO via live modification/tuning of the partitions as samples from the system are collected.} The proposed parallel paradigms can also open the door to a number of applications and potentially other decomposition paradigms that we will aim to explore in the future. Specifically, we are interested in exploring more complex systems that involve higher-dimensional design spaces and large numbers of parallel experiments. This will allow us to investigate the asymptotic properties of the proposed approaches. Moreover, we are interested in designing alternative paradigms that selectively exchange information between partitions to accelerate the search and that use different types of reference models to guide the search. We are also interested in exploring the application of these approaches to the tuning of complex controllers.

%%%%%%%%%%%%%%%%%%%%%%%%%%%%%%%%%%%%%%%%%%

\section*{Acknowledgments}

We acknowledge financial support  via the NSF-EFRI award 2132036 and the Advanced Opportunity Fellowship from the University of Wisconsin-Madison Graduate Engineering Research Scholars program.

%%%%%%%%%%%%%%%%%%%%%%%%%%%%%%%%%%%%%%%%%%

\appendix

\section{Reactor System Model}

\subsection{Exact Model}

The reactor system consists of a pair of CSTRs operating at steady-state and connected in series. In the first reactor, reactant $A$ is converted into a desired product $P$, which can react further to form an undesired product $U$. An additional reactant $D$  reacts with $U$ to form $A$, and is fed to the first reactor to reduce the amount of $U$ formed. In order to further reduce the amount of $U$ present and increase the value of the product stream, the outlet of the first reactor is then fed to a second reactor along with an additional reactant $B$, which can react with $U$ to form a secondary product $E$. The reaction mechanism is complex and given by:
\begin{subequations}\label{rxn_net}
    \begin{gather}
        2A\longleftrightarrow P\\
        P\longleftrightarrow 2U\\
        U+B\longleftrightarrow E\\
        U+D\rightarrow 2A
    \end{gather}
\end{subequations}
The rates of each reaction are assumed to be elementary and thus:
\begin{subequations}\label{rxn_exps}
    \begin{gather}
        r_1=k_1C_A^2-k_{1r}C_P\\
        r_2=k_2C_P-k_{2r}C_U^2\\
        r_3=k_3C_UC_B-k_{3r}C_E\\
        r_4=k_4C_UC_D
    \end{gather}
\end{subequations}
where $C_i$ is the concentration of species $i$ and $k_j$ and $k_{jr}$ are the forward and reverse rate constants of the $j^{th}$ reaction. In our analysis we assumed that $k_{jr}=0.01k_j$, indicating that the forward reaction is favored. The material balances are:
\begin{subequations}\label{mass_bal}
    \begin{gather}
        0=F_{in}C_{A_{in}}-F_{out}C_{A}-2(r_1-r_4)V\\
        0=F_{in}C_{P_{in}}-F_{out}C_{P}+(r_1-r_2)V\\
        0=F_{in}C_{U_{in}}-F_{out}C_{U}+(2r_2-r_3-r_4)V\\
        0=F_{in}C_{B_{in}}-F_{out}C_{B}-r_3V\\
        0=F_{in}C_{E_{in}}-F_{out}C_{E}+r_3V\\
        0=F_{in}C_{D_{in}}-F_{out}C_{D}-r_4V
    \end{gather}
\end{subequations}
where $F_{in}$ and $F_{out}$ are the volumetric flowrates of the reactor feed and outlet respectively, $C_{i_{in}}$ is the concentration of species $i$ in the feed stream, and $V$ is the volume of the CSTR. The reactions in \eqref{rxn_net} are assumed to be exothermic and a cooling jacket is used to remove excess heat and control the temperature inside of the reactors. The jacket uses a fluid entering at a temperature $T_{ic}$ and flowing at a mass flowrate of $\dot{m}_{c}$ as the coolant. The coolant flowrate required to maintain the desired temperature can be determined from the reactor energy balance:
\begin{subequations}\label{energy_bal}
    \begin{gather}
        H_{in} = \rho C_pFT_{in}\\
        H_{out} = \rho C_pFT\\
        \dot{Q}=-r_1V\Delta H_1-r_2V\Delta H_2-r_3V\Delta H_3-r_4V\Delta H_4\\
        \dot{m}_{c} = \frac{H_{in}-H_{out}+\dot{Q}}{C_{pc}(T_{oc}-T_{ic})} = \frac{\rho C_{p}F(T_{in}-T)+\dot{Q}}{C_{pc}(T_{oc}-T_{ic})}
    \end{gather}
\end{subequations}
where $T_{in}$, is the temperature of the inlet stream, $\Delta H_j$ is the heat of reaction for the $j^{th}$ reaction, and $C_{pc}$ is the specific heat capacity of the coolant. Additionally, we assume that reactions do not change the heat capacity $C_{pin}$ or density $\rho_{in}$ of the reactor inlet. This allows us to set $C_p = C_{pin} = C_{pout}$, $\rho=\rho_{in}=\rho_{out}$, and $F = F_{in} = F_{out}$. The relation between the rate constants and temperature is described by the Arrhenius equation:
\begin{equation}\label{Arrhenius}
    k = k_0\exp\left({\frac{-E_A}{RT}}\right)
\end{equation}
where $k_0$ is the pre-exponential factor, $E_A$ is the activation energy of the reaction, and $R$ is the universal gas constant. 
\\

The outlet of the second reactor is fed to series of flash separation units to recover the products from the effluent stream. Product $E$ is recovered in the vapor fraction of the first vessel as stream $\dot{v}_1$, and product $P$ is recovered in the liquid fraction of the second vessel as stream $\dot{l}_2$. The relative volatility of the chemicals is set with respect to the vapor-liquid equilibrium ratio $K_P$; compositions and flows for the exiting streams can be determined from the following vapor-liquid equilibrium calculation:
\begin{subequations}\label{flash_mass_bal}
    \begin{gather}
	x_i=\frac{z_i}{f\left(K_P\alpha_i-1\right)+1}\\
	y_i=K_p\alpha_ix_i
    \end{gather}
\end{subequations}
where $z_i$, $x_i$, and $y_i$ are the molar fractions of species $i$ in the feed, liquid, and vapor streams respectively. The relative volatility of each chemical is denoted by $\alpha_i$ and $f$ is the fraction (on a molar basis) of the feed that exits the vessel in the vapor stream. We set $f$ according to the molar fraction of the recovered product in the feed, $f=z_E$ for the first vessel, and $f=1-z_P$ for the second vessel. The energy required to vaporize the desired fraction of the flash's feed was supplied by a heater that uses steam as the heating agent. The required flowrate of steam, $\dot{m}_{stm}$, was determined from the flash vessel energy balance
\begin{subequations}\label{flash_q_bal}
    \begin{gather}
	\dot{Q}_{vap} = \sum_{i\in\{A, P, U, B, E, D\}}L_{i}y_{i}\dot{v}\\
	\dot{m}_{stm} = \frac{\dot{Q}_{vap}}{L_{H_2O}}
    \end{gather}
\end{subequations}
where $L_i$ and $L_{H_2O}$ are the latent molar heat of species $i$ and water, respectively.
\\

The performance of the system is expressed as a cost function (negative profit) that measures the quality of the product streams along with the corresponding utility requirements at various temperatures and is formulated as:
\begin{subequations}\label{f_ov}
    \begin{gather}
	f_{1}(T_1, T_2) = \sum_{i\in\{A, P, U, B, E, D\}}w_iy_{i1}\dot{v}_1+\sum_{i\in\{A, P, U, B, E, D\}}w_ix_{i2}\dot{l}_2+\sum_{i\in\{A, B, D\}}w_{i}F_{i}C_{i0}\label{prod_cost}\\
	f_{2}(T_1, T_2) = w_c\left(\dot{m}_c+\dot{m}^{\prime}_c\right)+w_{stm}\left(\dot{m}_{stm}+\dot{m}^{\prime}_{stm}\right)\\
	f(T_1, T_2) = f_{1}(T_1, T_2)+f_{2}(T_1, T_2)
    \end{gather}
\end{subequations}
where $T_1$ and $T_2$ are the operating temperatures of the first and second reactor. The molar fraction of species $i$ in first product stream, $\dot{v}_1$, is denoted by $y_{i1}$, and $x_{i2}$ is the molar fraction of species $i$ in the second product stream $\dot{l}_2$. The price of species $i$ is represented by $w_i$, and $w_{c}$ and $w_{stm}$ are the costs of the cooling and heating utilities respectively. The cost of the reagents supplied to the network is captured by the final term in \eqref{prod_cost} where $F_i$ and $C_{i0}$ are the volumetric flow rate and inlet concentration respectively of species $i$ into the process. 

\begin{figure}[!htp]
	\begin{center}
	    \begin{subfigure}
	        \centering
	        \includegraphics[width=6in]{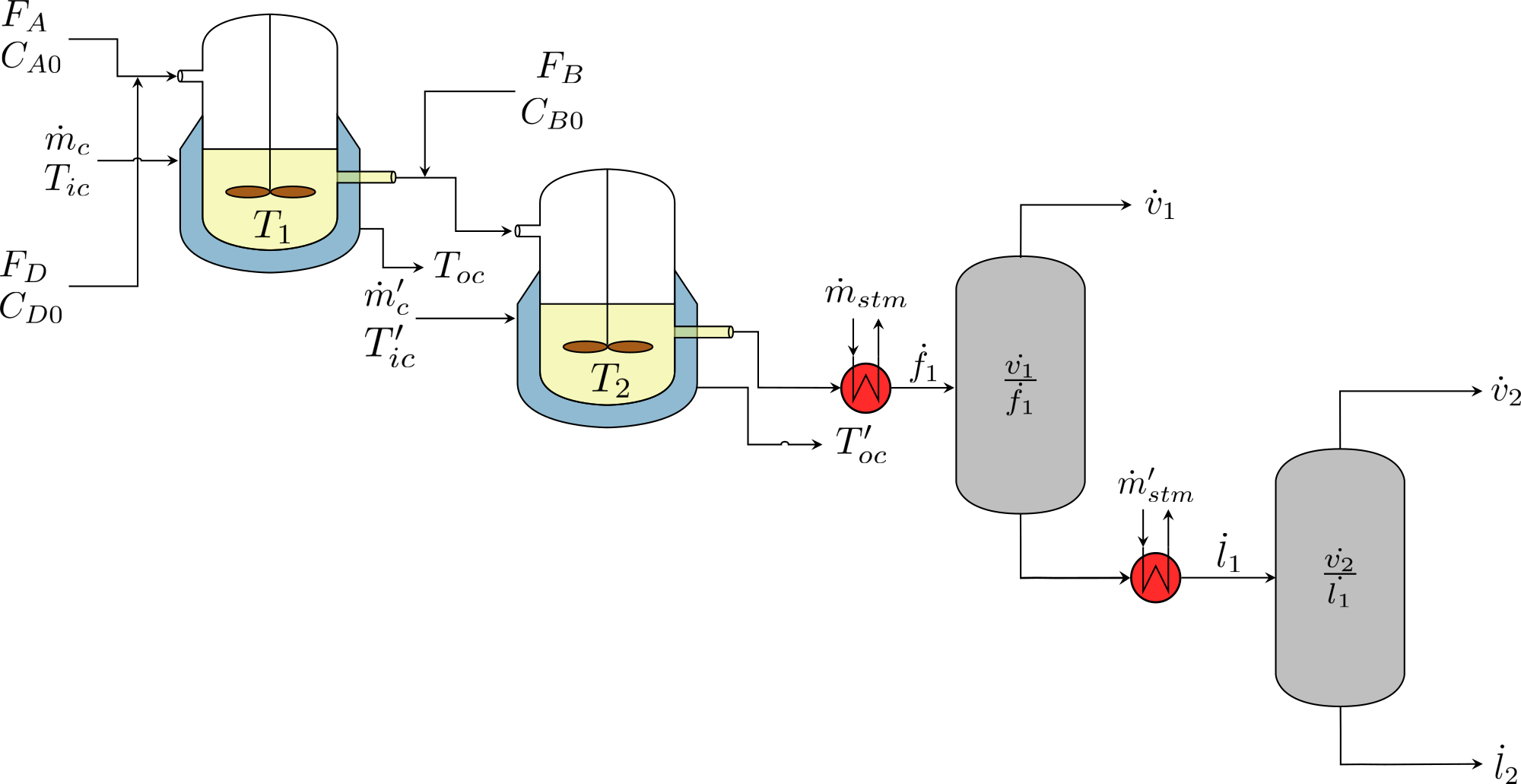}
	    \end{subfigure}
	    \caption{Schematic diagram of the serial CSTR reactor system and product recovery system}
	    \label{fig:rxtr_diag}
	\end{center}
\end{figure}

\subsection{Reference Model}

By substituting the Arrhenius expression \eqref{Arrhenius} into the rate expressions \eqref{rxn_exps}, we can determine that reaction rates are functions of temperature and concentration; this is a major source of nonlinearity in the system. We draw inspiration from the use of inferential sensors that are used in industry to correlate the rates directly to temperature (bypassing concentrations) in order to develop a reference model. Specifically, in our reference model we develop a polynomial function that approximates the dependence of the rate on the temperature. We develop our polynomial model based on the following transformation of the rate expression:
\begin{equation}
        \log{r}=\frac{-E_A}{R}\frac{1}{T}+n\log{C}
\end{equation}
where we ignore the reverse reaction due to the comparatively small $k_r$ values used. From the mass balances, we can also determine that the concentration is an implicit function of the temperature. We choose to capture this relation using higher-order polynomials. During our analysis, we determined that a third-order polynomial provided satisfactory performance, resulting in the following approximation for the rate expression:
\begin{equation}\label{rxn_approx}
    \log{r}=\theta_1\left(\frac{1}{T}\right)+\theta_2\left(\frac{1}{T}\right)^2+\theta_3\left(\frac{1}{T}\right)^3+\theta_0
\end{equation}
where $\theta_0$, $\theta_1$, $\theta_2$, and $\theta_3$ are the model coefficients. Using \eqref{rxn_approx}, we can rewrite the material balances purely as functions of temperature and obtain the following expressions for the various species concentrations:
\begin{subequations}\label{mass_bal_ref}
    \begin{gather}
        C_{A}=C_{A_{in}}-2(r_1(T)-r_4(T))\frac{V}{F}\\
        C_{P}=C_{P_{in}}+(r_1(T)-r_2(T))\frac{V}{F}\\
        C_{U}=C_{U_{in}}+(2r_2(T)-r_3(T)-r_4(T))\frac{V}{F}\\
        C_{B}=C_{B_{in}}-r_3(T)\frac{V}{F}\\
        C_{E}=C_{E_{in}}+r_3(T)\frac{V}{F}\\
        C_{D}=C_{D_{in}}-r_4(T)\frac{V}{F}
    \end{gather}
\end{subequations}
we then substitute these values into a performance function similar to \eqref{f_ov} to obtain the reference model for the system $g$.  This reference model can be seen as an approximate physical model of the real system (captured by the exact model) and is much easier to evaluate. However, because this model is comprised of a complex set of algebraic equations, we further approximate the dependence of the performance function on the temperatures using a GP model. 

\bibliographystyle{abbrv}
\bibliography{./vzavala}

\end{document}